\documentclass[twoside,11pt]{article}
\usepackage{jmlr2e}

\usepackage[utf8]{inputenc}

\usepackage{algorithm}
\usepackage[noend]{algpseudocode}

\usepackage{graphicx}
\graphicspath{ {./} }
\usepackage{mathtools}
\usepackage{hyperref}

\jmlrheading{}{2021}{}{8/21}{}{robinson}{Christopher Robinson}

\ShortHeadings{Goal Agnostic Planning on Max Likelihood Hypergraphs}{Robinson}

\firstpageno{1}

\begin{document}

\title{Goal Agnostic Planning using Maximum Likelihood Paths in Hypergraph World Models}

\author{\name Christopher Robinson \email ckevinr@gmail.com \\
       \addr Department of Electrical and Computer Engineering\\
       University of Louisville\\
       Louisville, KY 40292, USA
       }

\editor{}

\maketitle

\begin{abstract}%
In this paper, we present a hypergraph--based machine learning algorithm, a datastructure--driven maintenance method, and a planning algorithm based on a probabilistic application of Dijkstra's algorithm. Together, these form a goal agnostic automated planning engine for an autonomous learning agent which incorporates beneficial properties of both classical Machine Learning and traditional Artificial Intelligence. We prove that the algorithm determines optimal solutions within the problem space, mathematically bound learning performance, and supply a mathematical model analyzing system state progression through time yielding explicit predictions for learning curves, goal achievement rates, and response to abstractions and uncertainty. To validate performance, we exhibit results from applying the agent to three archetypal planning problems, including composite hierarchical domains, and highlight empirical findings which illustrate properties elucidated in the analysis. 
\end{abstract}


\section{Introduction}
Traditionally, AI planning and Machine Learning are conceptualized as solving somewhat different problems. Machine Learning generally casts problems as some form of pattern identification, and seeks to identify structural relationships which render problem solving trivial, while AI leverages relationships between states and attempts to apply logical operations to convert pattern--weak representations into solutions. This is a quite broad generalization, but in terms of trends the two fields often conform to these patterns. Herein, we seek to present a comprehensively designed system which is, at its heart, a method for turning learned representations of state spaces into solutions within those spaces. It applies components of classical machine learning algorithms (reinforcement learning in particular) to develop an empirical world model relating observed states, actions, and the results of those actions. These relations are maintained in a carefully arranged data structure on which classical planning algorithms (in this case Dijkstra's algorithm operating on probability) can extract a maximum likelihood plan for goal state achievement.

The Goal Agnostic Planner (or GAP algorithm), presented here, is thus a combination of a composite data structure, a simple algorithm which maintains this data structure, and an application of Dijkstra's algorithm to produce solutions. Combined, these synthesize into a learning--based system which is capable of determining action policies for problems without reliance on the design of reward or fitness functions. Additionally, the careful choice of structure for the learned world model means it is possible to analyze the system to establish guarantees for convergence, system dynamics, and tolerance to uncertainty and abstraction. Doing so, this method combines elements of reinforcement learning, Markov Decision Processes, and automated planning.

\subsection{Related work}

In this section we discuss the aforementioned learning and problem solving fields, with an eye towards highlighting similarities with our work and illustrating the typical limitations of these methods which we wish to ameliorate.

\subsubsection{Reinforcement Learning}

Reinforcement Learning (RL) is one of the earliest developed, and most widely used, machine learning models, as attested to by \cite{kaelbling1996reinforcement}, and can readily be applied to determining a wide range of action policies. RL agents operate in a state/action framework, and learn a form of objective function for maximizing prospective future rewards by taking certain actions based on observed state of the world. The use of the state/action framework as a broad approach to modeling systems has allowed for the development of many variants of reinforcement learning which aim to solve various problems. For instance, Temporal Difference learning by \cite{sutton1987temporal}, SARSA, developed by \cite{rummery1994line} and the classic Q-Learning by \cite{watkins1989learning}.

What this diversity of applications shows is that the state/action modeling framework is extremely effective for problem solving, and a common through--line for reinforcement--based systems is the effectiveness of reward assignment in satisfying convergence conditions, via the Bellman equation, canonically expressed in the proof of convergence for Q-Learning by \cite{watkins1992q}.

As such, the reward function is usually a critical point for the RL system and performance is predicated on the quality of this function, a relationship explored in detail by \cite{matignon2006reward}. However, one of the main limitations of reinforcement learning, expressed both by \cite{koenig1996effect} and by \cite{grzes2017reward}, is goal orientation. Reward functions are, by necessity, constructed in relation to a specific objective set, which makes training of the agent specific to the model scenario and the explicit goals implicitly or explicitly expressed by the reward function. 

While it is sometimes possible to transfer RL training from one problem case to another, provided they are sufficiently similar (a range of conditions denoting 'similar' have been identified by \cite{taylor2009transfer}), it is an entirely different matter to use the quality function to solve an already learned problem for an alternate goal state. 

Of particular note is that in learning the value of the reward function, not only is a designed parameter which is fixed with relation to the goal being learned, but also additional observational data which may be pertinent to future operations is lost. With our method, we seek to store and apply the learned \textit{state relationships} in a way which can be applied to problem solving between any pair of reachable states, preserving all such relationships for later use. We thus eliminate both goal dependence and information loss by extension of the state/action framework into a higher dimensional state/action/state space. 

\subsubsection{Markov Decision Processes}

Markov Decision processes represent an analytic approach to constructing agents which estimate the behavior of systems under uncertainty, and enable predictions of the system state under time evolution. \cite{white1985real} discusses, from a relatively early standpoint, some of the diversity of cases in which MDPs have found use in real world applications, speaking to their power as modeling tools. \cite{van2012reinforcement} even go so far as to discuss MDPs as the 'de facto' standard for sequential learning, a position that would be difficult to dispute in any but the most narrow fields of research.

One of the greatest utilities of MDPs is that they provide mathematical tools needed for optimization of semi--random processes under a state/action framework. As an example, \cite{ding2014optimal} present a method using MDPs for solving optimal controls problems, an interesting case showing how a continuous system may be examined using the inherently discrete system states for an MDP. MDPs have been also been used for optimizing planning processes, \cite{karami2009partially} show an application which schedules manual and autonomous activity in a collaborative robotics task, \cite{fox2001agent} approaches policy planning of material acquisition under an agent based framework, and \cite{floriano2019planning} showcase a fascinating use of MDPs for managing interaction in a fleet of UAVs. These highly varied cases highlight the potential for MDPs as means of analyzing the dynamics of complex systems.

While generally considered more flexible than reinforcement learning and other machine learning systems, MDPs are still typically reliant on the modeling of the system in question. Further, the reward function is critical for convergence of algorithms which use dynamic programming, via the Bellman equation, to solve for optimal MDP policies. For instance, \cite{dimitrov2009combinatorial} discuss the construction of action sets for MDP formulation as a design methodology, illustrating the presence of optimality conditions embedded within problem structure. In some cases, such as investigated by \cite{szepesvari1996generalized}, reinforcement learning has been used to supplant knowledge of the reward function, replacing it with an implicitly learned quality function. The indication here is that the representativeness of the reward function (here vis-a-vis reinforcement learning) is thus critical to productive application of MDPs.

In our analysis, we will rely heavily on Markov process analysis techniques to evaluate the performance of the algorithm and demonstrate learning convergence and robustness. To do so, we are using the fact that the GAP  algorithm's time evolution is readily modeled with an MDP, and the predictive power thereof allows us to define the temporal behavior of the system. The core difference, however, is that we define an optimal action policy initially, in the form of the \textit{maximal likelihood path}. This policy possesses useful properties for the Markov analysis, essentially converting the predictive power of the MDP into a problem solution without recourse to iterative policy estimates.

\subsubsection{Automated Planning \& Learning}

Automated planning is one of the oldest fields associated with artificial intelligence, and adaptive methods of applying planning algorithms have been investigated since the field's inception. Typically, planning algorithms are designed to solve problems which are inherently hierarchical and discrete, and implemented as combinatorial or dynamic programming algorithms, with even fundamental search and sort algorithms often playing a central role in planning.

Potentially the most iconic problem solving system, constructed with the aim of achieving robotic task accomplishment, was proposed by \cite{fikes1971strips},  the Stanford Research Institute Problem Solver, or STRIPS. The widespread effectiveness of STRIPS set the stage for many later problem solving systems, and \cite{lekavy2007expressivity} indicate that STRIPS is computationally complete within properly formatted world spaces, vindicating this general adoption.

There are many successors to STRIPS which use similar, semantically based logic systems. The key feature linking these disparate approaches is the use of semantic relationships to identify solutions, fundamentally casting the solution generation process as a search problem.

One of the primary such competing methods is the Hierarchical Task Network, originally outlined by \cite{sacerdoti1975nonlinear} and with a similarly large number of distinct implementations to STRIPS. HTNs are noted by \cite{georgievski2014overview} as possessing many of the same limitations as STRIPS, but have been generally considered more expressive of problem domains at the expense of requiring highly detailed models. Further, \cite{lekavy2007expressivity} demonstrate that both HTN and STRIPS are functionally equivalent in practical application, and hence we treat STRIPS as generally representative here.

Another, related, leading standard for planning is the Problem Domain Description Language, initially proposed by \cite{mcdermott20001998} and later extended by \cite{fox2003pddl2} to a broader system that can be applied to more problems, a response to growing usage of the language. Further development was made by \cite{younes2004ppddl1} to incorporate explicitly probabilistic effects of actions. These systems are oriented around semantic planning, based on building joint logical operations around objects with known properties which interact with the operators (or actions). They thus rely explicitly on world model design and rule allocation, expressly limiting scalability.

Work regarding these STRIPS and STRIPS-like systems has been ongoing for decades, especially via its use as an expressive language for programming AI problem solving applications. \cite{geffner2000functional} proposed one such expansion, a system for increasing the domain representativeness of world models. Retrospectively, \cite{lekavy2007expressivity} indicates that this is not strictly necessary for generalized application of the STRIPS algorithm, but it speaks directly to the import of human effort in design of world models, a feature we seek to remove from the process.

\cite{sacerdoti1974planning} presented a modification of STRIPS which can develop and operate within an abstracted world model of the same kind as the STRIPS algorithm, and as a result generate substantial improvements in planning capacity, illustrating the criticality of the use of abstraction models for solving complex problems. While this model inherits many of the limitations of STRIPS previously mentioned, it does illustrate that modifying the world model can have a substantial impact on performance. \cite{dicken2010applying} further develop on the use of clustering as a means of improving time performance, further bolstering this observation.

Efforts towards the use of learning to supplement modeling have been researched as well. \cite{zimmerman2003learning} and \cite{jimenez2012review} discuss a wide range of algorithms which apply learning methodologies to inform the operation of classical planning systems. \cite{jimenez2012review} elucidate two major issues within the field of automated planning which are well addressed by learning: (1) 'Accurate descriptions of learning tasks', and (2) 'failure to scale up or yield good quality solutions'. \cite{bylander1996probabilistic} in particular, though writing earlier, addressed several bounding propositions related to problem (2) in propositional STRIPS-like planning, similar to those we will be developing herein, but without the benefit of relationships derived from the hypergraph data structures we present.

\cite{blum1997fast} present Graphplan, which performs planning operations on a task graph representing a hierarchical domain world model, implicitly an effort towards improvements over problem (2). Graphplan presents a means of encoding information about the problem structure into a simpler model, and then planning on that new model. The author notes that this allows for much more tractable planning speeds which are polynomial time bounded,  fundamentally a feature of changing the objective of the planning algorithm from a search problem, as in semantic planners, to that of a path finding approach. The authors later extend their work to probabilistic planning (\cite{blum1999probabilistic}), however, they directly acknowledge limitations associated with multiple overlapping action results (a special case of problem (1) above), an issue we resolve using the hypergraph learning structure.

Similarly, \cite{leonetti2016synthesis} combine reinforcement learning with automated planning, implementing their DARLING algorithm, using the learning system to identify reduced sized MDPs representing the problem space. With this approach, they are able to identify conditional resilience to uncertainties, and observe performance improvements under learning. However, their method still relies on the definition of a reward function, and does not incorporate performance improvements associated with graph based planning. Where Graphplan incorporates  non-search planning methods, but is limited by the construction of the graph from propositions and does not use learning, DARLING implements learning and reduces the size of the problem space, but still is limited by the reward function definition and semantic search methods.

We address the problems illustrated above by modeling the planning task as a lower order combinatorial problem operating on a datastructure which allows for in--situ maintenance of the critical relationships between states, confining the space complexity to $O(n^3)$, and the time complexity to $O(n^2)$ using a standard implementation of Dijkstra's Algorithm (\cite{dijkstra1959note}), rather than semantic logic. This lets us address issue (1) by treating the learning task as a probability optimizing planning path, and issue (2) by retaining all the functional operations well within polynomial time space, combining aspects of prior work approaching these outstanding problems in a coherent single system.

\subsection{Contribution}

As we have seen, a substantial limitation of machine learning systems is the need for the determination of objective functions to drive convergence of learning to an optima. Generally speaking, these functions serve at a very high level to convert a complex non-linear (possibly even combinatorial) workspace into a reasonably convex space on which various optimization methods can be applied. Typically, this requires certain constraints on these functions, and on basis of those constraints are derived performance guarantees by analysis via the Bellman Equation, such as described by \cite{vidyasagar2020recent}. \cite{baird1999gradient} presents a similar argument, following the use of Gradient Descent, as another example which illustrates the point of the optimization argument.

By contrast, MDPs and planning algorithms typically rely on the quality of the world description, a property expressly investigated in \cite{dimitrov2009combinatorial}, where the authors explore the problem of optimal design of action sets for MDPs, very intimately related to the optimal policy choice embedded in our method. Additionally, the use of heuristics or modeling restrictions for reduction of the problem space to tractable sizes, as seen from \cite{dicken2010applying}, has been investigated towards the end of improving model representativeness while maintaining efficiently computable space sizes. While success has been seen with these methods, there are still inherent limits imposed by the world construction and design of space reducing rules. Often, this exposes the learning system to designer bias, or loss of information. While automated planners rarely include implicit goal dependence, they do not generally provide for direct adaptation and expansion of their state spaces, detailed as a primary challenge both early by \cite{sacerdoti1974planning} and substantially later still by \cite{jimenez2012review}.

To address these limitations specifically, this paper describes a composite data structure along with maintenance and inference algorithms which together comprise a learning system. Further, we present an analysis of the resulting system which provides bounded stochastic guarantees on convergence performance, planning efficacy and optimality. We are also able to use this analysis to discuss the impact of uncertainty and abstractions on the performance of the system. By implementing a holistic approach to management of the learned data and the planning algorithm, we are able to derive several measurable benefits:

I. Goal agnosticism- By design, any state can be set as the goal and the native probability vector replaced with an associated 'goal state probability vector', and system dynamics are preserved.

II. Algorithm efficacy- We present a proof that the algorithm extracts the optimal path to the goal, even under stochastic uncertainty.

III. Abstraction robustness- We present mathematical proof of convergence under an abstraction, modeled as an unknown state space transform, and a derived metric that can be used to represent the quality of the mapping, which has implications towards error tolerance as well.

IV. Learning Convergence- We present both a proof that the system's learning converges, and a derivation of the expected learning curve in the average case over many instances.

Additionally, we investigate three problem domains in detail which demonstrate adherence of actual performance to these properties predicted by the analysis, and further explore relationships among the properties by evaluation of the experimental data.

The remainder of the paper is organized as follows: in the next section, we define the fundamental components of the system as we will be using them throughout the paper- some features are common to extant systems, and some are distinct from them. Following that, in Section 3 we present the algorithms and data structures which comprise the GAP algorithm. Section 4 includes the analysis of the algorithm, including proofs of efficacy, the dynamic behavior of agents implementing GAP, analysis of the effect of abstractions on performance, and finally demonstration of learning convergence. In Section 5, we present our demonstration cases and experimental procedures, data collected from these experiments, and discussions thereof in the context of the prior analysis. Finally, in Section 6 we conclude the paper and discuss avenues of future research.

[ENDED EDITS HERE]

\section{Definitions}

Herein, we define components of the modeling strategy which we use to effect comprehensive learning of the state space that can be leveraged by the GAP algorithm.

\subsection{Agent \& Occasions}

The agent is presumed to be the portion of the system capable of making decisions and effecting the world. It is defined by the capability to register a set of perceptual \textit{states} (denoted $S$) and take a set of \textit{actions} ($A$), which can impact the world and possibly alter the state. At any given point in time $k$, the agent can observe an initial state, $s_i$, and subsequently take an action $a_l$, resulting in a state change to a final state $s_f$ (note that $s_f$ may be identical to $s_i$). Such a series is henceforth referred to as an \textit{occasion}: $o_k = a_l(s_i)\rightarrow s_f$. Occasions are the atomic unit of the algorithm's data retention, described shortly.

\subsection{Hypergraph Learning Model}

\begin{figure}[t]
\centering
\includegraphics[scale = 0.75]{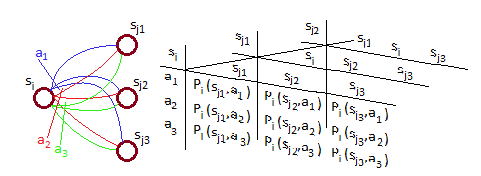}
\caption{Hypergraph representation in 3d array, detailing the existence of multiple overlapping edges between pairs of nodes for which differing actions may result in the same state to state transition.}
\label{fig:Hypergraph}
\end{figure}

We implement a learning system for which the basic units are occasions, defined in the prior section, and are recorded within a 3-dimensional structure of size $|S|\times|S|\times|A|$. Within this array, cells at locations $(s_i,s_f,a_l)$ contain an instance count of the number of times the corresponding occasion has been observed. This data structure is conceptualized as a directed hypergraph:  a higher dimensional analog of a graph in which each state is a node and initial and final state labels express edges corresponding to out edges and in edges respectively. For each node, then, the dimensional expansion results in multiple edges between each node, corresponding to varying actions. Each action may have multiple results, and these results may overlap with other actions' results and thus we have multiple links between states, each connected with a different action, hence the construction of a hypergraph as opposed to a standard graph.

\begin{figure}[t]
\centering
\includegraphics[scale = 0.65]{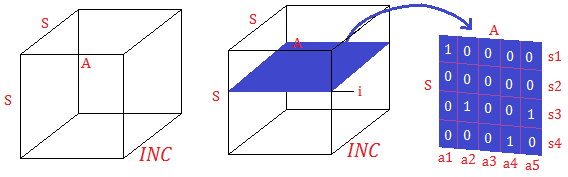}
\caption{Extraction of an Action/State results slice from the hypergraph, illustrating the relationship between the a priori state $s_i$ and the potential results of taking actions from within that state.}
\label{fig:SA_slice}
\end{figure}

To represent the hypergraph in practice, we implement a three dimensional array, INC, in which the location $INC[i,j,l]$ contains the number of times occasion $a_l(s_i)\rightarrow s_j$ has been observed. An illustrated reference is presented in \autoref{fig:Hypergraph}. Using the member entries of the INC array, and sums along slices within this array (as illustrated in Figure \ref{fig:SA_slice}), the relative probability of differing occasions can be computed.

\subsection{Probability Models}

As mentioned above, we can calculate the probability of an occasion occurring as a ratio of the number of observed instances of the occasion to the sum of occasion counts along a slice of INC. However, as $s_i$ is fixed but $a_l$ and $s_f$ are not, this presents two possibilities for probability models, one referenced against resultant states and one referenced against actions taken. 

In the first model, we elect to choose actions based on the most probable \textit{outcome} of taking an action from a given state. Calculation of the associated probabilities is thus given by the following formula:

\begin{equation}
P(a_l(s_i)\rightarrow s_j) := P(s_j|s_i,a_l) = \frac{INC[s_i,s_j,a_l]}{\sum_{\forall s } INC[s_i,s,a_l] }
\label{eqn:apri_prob}
\end{equation}

This approach conceptualizes each action in terms of which state is most likely to be observed next as a result of taking that action, and hence we refer to it as the \textit{a priori} probability.

In the second model, we select actions based on the most probable cause of a given state/state transition, which we term the \textit{a posteriori} probability. That is to say, we select, for each $s_i \rightarrow s_j$, the probability of the action $a_l$ having caused this transition relative to other actions. Therefor, in this model, the probability calculation is performed as:

\begin{equation}
P(a_l(s_i)\rightarrow s_j) :=  P(a_l|s_i,s_j) = \frac{INC[s_i,s_j,a_l]}{\sum_{\forall a } INC[s_i,s_j,a] }
\label{eqn:apost_prob}
\end{equation}

We can see the difference, for instance, if we presume a simple problem with three states and three actions, with an INC slice at $s_i$ as demonstrated in Table 1:

\begin{table}[H]{
\centering
\begin{tabular}{|c|c|c|c|}
\hline
$s_i$        & $a_1$ & $a_2$  & $a_3$ \\
\hline
$s_{i}$     & 3 & 1 & 7 \\ 
$s_{f1}$    & 2 & 5 & 1 \\
$s_{f2}$    & 9 & 1 & 2 \\
\hline
\end{tabular}
\caption{Illustrative example of a state/action slice}
}\end{table}

We can see that Equation \ref{eqn:apri_prob}, when applied to $o_{k} = a_1(s_i)\rightarrow s_{f1}$, gives a probability of $\frac{1}{7}$, whereas Equation \ref{eqn:apost_prob} produces $\frac{1}{4}$. The former tells us that there is a 1-in-7 chance that taking $a_1$ from $s_i$ will result in $s_{f1}$, whereas the latter expresses a 1-in-4 chance that the transition $s_i \rightarrow s_{f1}$ can be precipitated $a_1$.

In the qualitative sense, then, action policies using the a priori probability model are selecting the sequence of actions most likely to result in goal achievement. This means that, in our policy consideration, each action will be a viable choice, with the most likely subsequent state as the primary outcome, even if multiple actions have the same most likely resultant state. Comparatively, a posteriori policies select the most likely sequence of state changes to reach the goal, meaning each policy choice will examine all possible state/state transitions from $s_i$ in terms of the highest probability action precipitating that transition, even if that action is the most likely cause of most available transitions.

\subsection{Sequences}

To effect a net change among  non-adjacent states, several actions must be taken. An ordered series of these transitions and the associated states we term a \textit{sequence}. Solutions produced by the planning algorithm are sequences, and maintenance of the action list which precipitates the transitions between states, and the states themselves, allow the execution of the solution by the agent. A sequence is represented as a pair of two ordered lists $\sigma_{og} = [\{s_1,s_{2}...s_g\},\{a_1,a_{2},...a_{g-1}\}]$, where $a_1(s_1) \rightarrow s_{2}$, $a_{2}(s_{2}) \rightarrow s_{3}$, and so on.

For each occasion we will have the conditional probability which represents the likelihood of the occasion occurring. For a sequence, we can then define a joint probability of the entire sequence being executed:

\begin{equation}
P(\sigma) = \prod_{\forall o_k \in \sigma}  P(a_l(s_i)\rightarrow s_j)
\label{eqn:net_prob}
\end{equation}

Throughout this paper we will be using this joint probability formula, which necessitates the assumption that occasions are conditionally independent. If the chosen states are sufficiently disjoint, this requirement is satisfied; if not, then the probabilities are presumed mixed in the same capacity as fundamental error, which we explore in Section 4.

For each occupied state, the potential effects of an action are represented in the current state slice of the hypergraph, as highlighted in \autoref{fig:SA_slice}. Each entry in $INC[s_i,:,:]$ therefor also represents an associated probability of relationship between action $a_l$ and the state transition $s_i \rightarrow s_j$, depending on which probability model is used.

\subsection{Maximal likelihood subgraph}

Given a specific probability calculation, and in the context of maximally likely sequences, it is possible to define a class of subgraphs embedded within the hypergraph which contain all component edges of a solution sequence. One such subgraph formulation which is computationally simple to construct and maintain contains all maximally probable transitions between any state pair $(s_i,s_j)$, stored as an $|S|\times|S|\times2$ array. In this array, the component $<s_i,s_j,0>$ is the maximum probability associated with the $s_i \rightarrow s_j$ transition, and component $<s_i,s_j,1>$ is the index of the corresponding action. Thus defined, we have:

$$AFI[s_i,s_j,0] = \frac{INC[i,j,\underset{l}{\operatorname{argmax}} \{ P(a_l(s_i)\rightarrow s_j) \}]}{\sum_{\forall s } INC[s_i,s,a_l] }$$

$$AFI[s_i,s_j,1] =  \underset{l}{\operatorname{argmax}} \{ P(a_l(s_i)\rightarrow s_j) \}$$

Another such graph, prepared and maintained similarly, is one which contains maximally likely final states with respect to actions taken. This graph can be represented on an $|S|\times|A|\times2$ sized array, in which members at $<s_i,a_l,0>$ represent the probability associated with the most likely result of taking action $a_l$ from state $s_i$, and $<s_i,a_l,1>$ represents the index of $s_f$. Note that, at this point, we have effectively defined the action choice policy for the system, which we shall later prove to be optimal.

The construction of each such subgraph from the larger hypergraph structure is accomplished simply by taking the projection of maximally probable elements from any directional slice of the hypergraph. For instance, to construct the A Posteriori subgraph, the hypergraph is compressed along the $|A|$ axis, retaining the maximally probable $a_l$ associated with each state transition in the $|S|\times|S|$ space. Maintenance of this maximal probability property is handled with the array/linked list datastructure as described below, and is presumed to proceed as the learning phase progresses, allowing for in situ sorting and efficient updates through the datastructure as each occasion observation is recorded.

Each of these compressed arrays can represent a traditional graph, which feature we will use for efficient computation of solution sequences. However, the two are not identical in structure, and consequently remit different associated probability calculations with comparably diverging interpretations, as described above.

\section{Datastructures \& Algorithms}

In this section, we begin by presenting the datastructures used to retain observed information, and then the algorithms which operate on these datastructures to identify solutions within the problem space.

\begin{figure}[t]
\centering
\includegraphics[scale = 0.5]{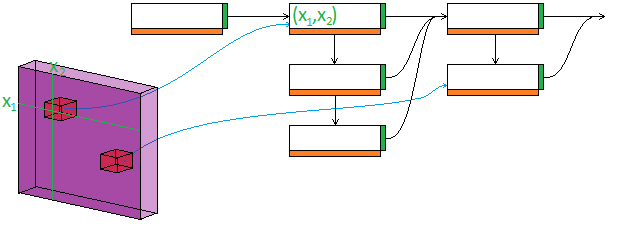}
\caption{Array/linked list, showing the indexed cell locations within the array containing pointers to the corresponding elements in the sorted linked list, which itself contains the data component associated with each array cell, and is organized into columns containing the same number of observed instances.}
\label{fig:ALL_diagram}
\end{figure}

\subsection{Array Linked List}

This datastructure combines an array with a linked list, as illustrated in \autoref{fig:ALL_diagram}, such that each element in the array is a pointer to a member of the linked list containing that address' necessary data. In such a structure, each array element contains a pointer to a member link within the linked list and each such member, in addition to any other data, contains its corresponding location within the array. 

In his way, the linked list need not be searched for member elements, and ordering of the list can be maintained using single operations on the linked list members. For our case, sorting is by incidence counts, and so we also implement the linked list in a parallel configuration with each 'column' containing instances with identical numbers of observed instances, so that each observation requires at most two operations to retain the list in sorted order.

\subsection{Augmented Hypergraph Datastructure}

\begin{figure}[b]
\centering
\includegraphics[scale = 0.6]{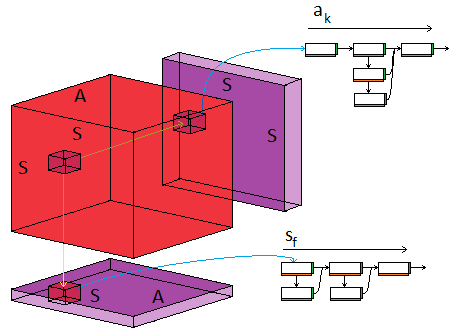}
\caption{Augmented hypergraph data structure: a 3 dimensional array, each cell of which contains pointers to members of two Array/Linked List objects, each containing a pointer to the corresponding sorted list associated with that state/action or state/state pair, allowing immediate retrieval.}
\label{fig:augmented_hypergraph}
\end{figure}

A hypergraph may be stored in a 3-dimensional array, and we retain the full $|S|\times|S|\times|A|$ sized collection of observation counts in such an array. For purposes of planning sequences, however, we must choose a probability model as described above. Under such a model, execution of the planning algorithm need not evaluate all hypergraph links, only the most probable ones, and thus for computational efficiency, the 3-dimensional array is augmented with a pair of array/linked lists corresponding to the sorted elements within the maximal likelihood subgraphs, illustrated in Figure \ref{fig:augmented_hypergraph}.

Using the convenient sorting and addressing features of the array/linked list, each update to the optimal subgraph can be incorporated using $O(1)$ alterations to the linked list containing the pertinent occasion. To accomplish this, the 3-dimensional array is paired with two 2-dimensional array linked lists, corresponding to the a priori and a posteriori probability metrics. Each member of these arrays points to a linked list which contains the sorted members of the slice along the compressed axis, and the graph associated with this compression consists solely of the first (and thus, maximally probable) element along that axis.

Figure \ref{fig:augmented_hypergraph} shows how each cell in the 3-dimensional array contains two pointers, one to each subgraph compression, and each subgraph cell contains the linked list of members along the compression axis. For the a priori probability, this corresponds to all cells with constant $(s_i,a_l)$, so the linked list contains $|S|$ members along the $s_j$ axis. Similarly, for the a posteriori graph, the linked lists are along constant $(s_i,s_j)$, and contain $|A|$ many links.

\subsection{Subgraph Maintenence Algorithm}

\begin{algorithm}[t]
\caption{Linked list subgraph maintenence}
\label{alg:LL_maint}
\begin{algorithmic}
\Function{MaintainLL}{$INC,(s_i,s_f,a_l)$}

\State $occasionLink = INC[s_i,s_f,a_l,0]$
\\

\If{$occasionLink.prev == None$}

return $1$

\EndIf

\If {$occasionLink.prev.count$

\indent\indent $> occasionLink.count$}

return $1$

\EndIf

\If {$occasionLink.prev.count$

\indent\indent $== occasionLink.count$}

\State $occasionLink.prev = occasionLink.prev.prev$
\State $occasionLink.post = occasionLink.current$
\State $occasionLink.current =$

\indent\indent$ occasionLink.prev.current$

\EndIf

\EndFunction
\end{algorithmic}
\end{algorithm}

To maintain the maximal likelihood subgraphs, at each observed instance of an occasion the linked lists containing references to this set of coordinates must be updated. Because the linked list members each contain increment counts of the number of times the occasion has been observed, and are sorted by these counts, each link may only move ahead one link in the list at any time. As such, maintenence revolves around correctly rebuilding the link chain at each step, as detailed in Algorithm \autoref{alg:LL_maint}.

Under the continual operation of this maintenance algorithm, the maximum likelihood subgraph slice of the bulk hypergraph data structure is perpetually embedded within the 2D slice corresponding to the first sorted members of each state/state or state/action list. Further, at each update step, the addressing to the linked list element is direct via the augmented hypergraph, and the only operations necessary are the direct comparison of the increment counts to the preceding linked objects, which on increment require only rebuilding  the links to the prior and current list members, and thus each update's complexity is $O(1)$.

\subsection{Sequence Inference Algorithm}

With the subgraph compression as described above in place, we are then in position to infer a maximal likely sequence of occasions to achieve a transition between the given, current, state $s_i$, and some other goal state, $s_g$. To identify the maximally probable path, we implement a modified version of Dijkstra's algorithm adapted to find maximum probability (rather than minimum weight) subtrees rooted at $s_i$ using the Array Linked Lists of AFI.

\begin{algorithm}[t]
\caption{Sequence inference algorithm}
\label{alg:seq_inf}
\begin{algorithmic}
\Function{SequenceInfer}{$AFI,(s_i,s_g)$}

\State $bound \gets s_i.edges$

\State $perm \gets [(s_i,1.0)]$

\State $edges \gets []$

\While{$s_g \notin permanent$}

    $jointProb(s_j) \coloneqq perm[bound[j]][1] \cdot{} bound[j].P$
    
    $s_{maxP} \gets \underset{s_j}{\operatorname{argmax}}(jointProb)$

    $perm \gets (s_{maxP},jointProb(s_{maxP}))$
    
    $bound = (bound \cup s_{maxP}.edges) - [e | e(1) = s_{maxP}]$
    
    $edges \gets bound[s_j]$

\EndWhile

$solution = [edges[s_g]]$

\While{$solution[-1][0] \neq s_i$}

    $solution \gets edges[solution[-1][0]]$

\EndWhile

\hspace{-0.15cm}\textbf{return} $solution$

\EndFunction
\end{algorithmic}
\end{algorithm}

In Algorithm \autoref{alg:seq_inf}, we define this algorithm calculating the net probability as a result of each sequential action. Additionally, we use the ALL data structure to make the ordering, member checking, and set operations of the boundary list simplistic and compatible with an implementation of Dijkstra's algorithm.

Dijkstra's algorithm typically searches through the minimum cumulative path sum through nodes within a growing adjacency list of a minimum path tree. In our case, the cumulative sum of distance for each node is instead the highest cumulative probability, joined under multiplication as expressed in Equation \ref{eqn:net_prob}. We will later present an explicit proof of the efficacy of this algorithm. Due to the structure of the augmented hypergraph, the computational efficiency of this method is $O(|S|^2)$ for the a priori probability model and $O(|S|\cdot|A|)$ for the a posteriori model, though other implementations could be used with somewhat improved characteristics, or A* if appropriate heuristics are known (discussed briefly in Appendix A.4).

\section{Algorithm Analysis}

In this section, we analyze the performance of the GAP algorithm as defined above to show optimality and efficacy, determine agent dynamics, the effects of abstractions on performance, and learning convergence properties.

\subsection{Effectiveness \& Optimality}

In implementing the subgraph compression, one may question whether or not there is information lost relative to the optimal path including all possible elements within the transition space. Here, we demonstrate that the optimal path is embedded within the subgraph.

\begin{theorem}
The solution with maximum joint probability within a hypergraph is embedded within the maximum likelihood subgraph.
\end{theorem}

\begin{proof}
We proceed by contradiction. Presume that there exists an optimal solution sequence $\sigma_{og}$ which contains an occasion not allocated to the maximum likelihood subtree. In this case, by definition the occasion must have an associated probability less than that of the corresponding transition in the subgraph. However, because probabilities are necessarily monotonically decreasing, the sequence $\sigma_{og}'$ using the subgraph's instance for the given transition will have higher probability than the assumed solution, and thus $\sigma_{og}$ is not optimal.
\end{proof}

Note that this proof applies to either probability calculation, as supplanting any element of a sequence is necessarily probabilistically related to the most probable member along the slice associated with that probability calculation.

Given that the optimal path is known to be embedded within the maximum likelihood subgraph, we then need to demonstrate that the inference algorithm is capable of extracting the solution.

\begin{theorem}
The sequence inference algorithm extracts the $\sigma_{og}$ representing the maximal joint probability sequence representing a path from $s_i$ to $s_g$.
\end{theorem}

\begin{proof}
Consider that all probabilities are on the range $[0,1]$, and that the joint probability function is therefor monotonically decreasing. We proceed by induction on the distance from $s_i$. The first node selected will have the maximum probability edge of all leading from $s_i$, and thus any alternate path to this node is bounded by that single probability. Continuing on, at any point in the sequence, each successive joint probability is further bound by the product of the prior and current occasion. As such, any higher probability bound occasion would have to be off of the maximum probability tree in AFI, a contradiction to Theorem 1.
\end{proof}

\subsection{Analysis of Agent Behavior Dynamics by Markov Chain}

Because we have been discussing probabilistically driven state/action transitions it is natural to make a comparison to Markov chains. In fact, it is possible to examine the behavior of the agent on a learned model as though following a Markov process by analysis of the maximal subgraph and the action selection policy implemented via Algorithm 2.

\subsubsection{Predictive Behavior Analysis}

To begin with, we have the maximal subgraph. For planning purposes we proceed by finding the highest probability expected path to the goal, but for this analysis, we do not have a specific starting state in mind. In lieu of this, we instead build the tree of maximal probability paths rooted in the goal state. We denote this tree as $T_{P(g)}$.

Within this tree, each maximally likely transition contains also the associated action most likely to effect that transition: the action the agent will choose when in that state. However, because each action is assumed to be non-deterministic there are multiple potential outcomes for the effectively fixed output choice across a range of probabilities: $a_l(s_i) : \{(s_{j1}|P_{j1}),(s_{j2}|P_{j2}),...\}$. 

\begin{figure}[t]
\centering
\includegraphics[scale = 0.7]{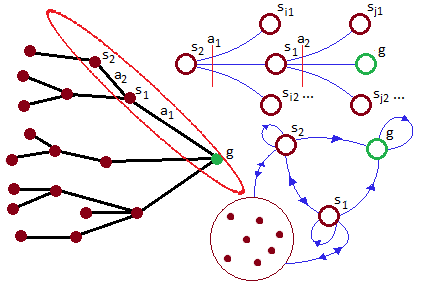}
\caption{Conversion of Maximal Probability path tree to Markov Chain. highlighting states $s_2$, $s_1$, and $g$, with maximal probability actions $a_2$ and $a_1$ linking them, and additional action ,based probability links among themselves and other states shown in the Markov network}
\label{fig:maxProb_to_markov}
\end{figure}

It is then possible to construct a Markov chain by constructing the transition table, selecting for each state the corresponding stochastic vector $AFI[s_i,a_{maxP},:]$ to the selected action for that state, excepting the goal state. This conversion is shown in representational format in \autoref{fig:maxProb_to_markov}. It is important to note here that the conversion does not result in a tree, actions are presumed stochastic, and thus any non-optimal transitions are also embedded in the Markov chain, so long as they result from \textit{optimal actions}.

For $s_g$ the operation of the agent effectively terminates, and so $\vec{s_g}$'s entries are 0 excepting the entry $\vec{s_g}[g]$, which is 1. From this concatenation, we have the transition table, $P_g$, which enables us to acquire a picture of the system behavior over time.

We can model the statistical propagation from a starting state by representing the state distribution itself as a vector, $\vec{s_k}$, where $k$ is taken to be an indicator of stepping, incremented each time an action is taken. At $k=0$, we have the known starting state at $P(s_0) = 1.0$, and so $\vec{s_0}$ is also a zero vector excepting the $s_0$ element, which is 1. The state occupation distribution as a function of step time is then easily given by:

\begin{equation}
\vec{s_k} = P_g^{k}\cdot{\vec{s_0}}
\label{eqn:state_prob}
\end{equation}

which represents the stochastic vector of probable states evolved from $s_0$ over time; and further that corresponding column of $P^k_g$ represents the probability of state occupation at step $k$ for the given starting state.

One thing to note about $P_g$ is that all columns represent probability vectors over states, such that $\Sigma_{j} P_g[i,j] = \vec{1}_{1\times|S|}$. Consequently, $||\vec{s_k}||_{1} = 1$, which is sensible as it is a probability vector. Now, because $\vec{s_k} =  P_g\cdot{}\vec{s_{k-1}} =  P_g(P_g^{k-1}\cdot{\vec{s_0}})$, we can define the stationary state distributions by $||\vec{s_k} - \vec{s_{k-1}}||_{1} < \epsilon$, or: $P_g\cdot{}\vec{s_{k-1}} \leq  (1 + \vec{\epsilon})\vec{s_{k-1}}$, which further implies that:

$$P_g^{k}\cdot{}\vec{s_{k-1}} \leq  (1 + \vec{\epsilon})^{k}\vec{s_{k-1}}$$
$$P_g^{k}\cdot{}\vec{s_{k-1}} \leq  (1 + \vec{\epsilon})^{k}P_g^{m}\vec{s_{k-m-1}}$$, thus:
$$P_g^{k-m+1}\cdot{}\vec{s_{k-1}} \leq  (1 + \vec{\epsilon})^{k}\vec{s_{0}}$$

Which necessitates that any attractor state \textit{either} be an eigenvector of $P_g$ with $\lambda = 1$, or be made arbitrarily close to an initial state vector. The $\lambda = 1$ eigenvector of any Markov transition matrix is also a steady state of the transition matrix. However, $P_g$ can be demonstrated to have no steady-state \textit{distribution} by virtue of the presence of $s_g$. Presume, w.l.g, that we order the stochastic vectors comprising $P_g$ such that $s_g$ corresponds to the last element. We then have:

$$P_g  = 
\begin{pmatrix}
T_s & \vec{0} \\
\vec{t_g} & 1\\
\end{pmatrix}
$$

Where $T_s$ is the transition matrix internal to only  non-goal states, $\vec{t_g}$ is the vector of transition probabilities from $\{s_i \in S | i \neq g\}$, and the final column is the stochastic vector of $s_g$, so structured as $<\vec{0},1>$ because we presume that execution terminates upon reaching the goal state. With this structure, we can then write:

\begin{equation}
P^k_g  = 
\begin{pmatrix}
T^k_s & \vec{0} \\
\vec{t_g}\cdot{}\Sigma^{k-1}_{l=1} T^l_s + \vec{t_g}& 1\\
\end{pmatrix}
\label{eqn:pg_timestep}
\end{equation}.

This expression precisely describes the probability distribution, as a function of step number, of the agent's occupation of states under the GAP algorithm. From it, we can see that the probability of reaching the goal state at step k is given by 

\begin{equation}
\vec{P}(s_i \rightarrow s_g|k) = \vec{t_g} \cdot{} \sum_{m=1}^{k-1} T^m_s + \vec{t_g}
\label{eqn:goal_vec}
\end{equation}

Which equation expressly describes the probability of any state transitioning to the goal state at a given step $k$. We can further note that for a given state distribution $\vec{s}_k$, at time $k$ we can express the probability of transition to the goal state at some future time $k' = k + \delta k$:

\begin{equation}
P(s_g|\vec{s}_k,k') = (\vec{t_g} \cdot{} \sum_{m=1}^{\delta k-1} T^m_s + \vec{t_g})\cdot(\vec{s}_k)
\label{eqn:goal_probs}
\end{equation}

Because $T_g$ is strictly positive definite, $T_g^k$ is as well, and consequently $P(s_i \rightarrow s_g|k)$ is monotonically increasing in $k$, and therefor $P_g$ has no steady state. This means that $s_g$ must, by definition, be an attractor state, as it is identical to its own start-state distribution. Second, this also means that no other state can be an attractor \textit{unless} there is a zero probability of transitioning out from that state. Such states may be present in $P_g$ due to the stochastic nature of $a_l$ possibly leading to states not on the maximal probability tree.

We can therefore not only predict the expected behavior of the agent, but also define the probability of reaching the goal state at any given step number, and thereby establish the expected number of steps to reach the goal. Perhaps even more importantly we can also, for non-goal states which are also attractors, calculate the probability of the agent being sequestered at said states by incidental variance.

Additionally, because the columns of $P_g$ are stochastic, we can also make the following relation:

$$\vec{1}_{1\times|S|} - \vec{1}_{1\times|S|}T_s^k = \vec{t_g}(\Sigma^{k-1}_{m=1} T^m_s + I_{|S|})$$

\begin{equation}
\vec{1}_{1\times|S|} = \vec{1}_{1\times|S|}T_s^k + \sum^{k-1}_{m=1} \vec{t_g}T^m_s + \vec{t_g}
\label{eqn:mag_conv}
\end{equation}

which bounds the progressive magnitude of any vector representing the probability distribution across the system states as a function of step time. Because we have demonstrated that there are no steady states embedded within $P_g$ barring those constructed in the same form as a goal state, $<\vec{0},1>$, and because $T_s^k$ is positive definite with all entries less than or equal to one, the probability distribution of goal transitions must be strictly monotonic over time.

\subsubsection{Trap nets}

Naturally, we have assumed the form of the goal state as $<\vec{0},1>$, which makes it an attractor state. We noted above that other states with this stochastic vector would also act as unwanted attractor states from which the agent cannot reach the goal state, and that otherwise no steady states exist. However, it is also perfectly possible for state sequences which are not independently attractor states, but which present no path to the goal once reached, to be non-steady attractors. Such a segments we will refer to as 'trap nets', collections of states from which the agent cannot proceed to the goal, such as illustrated in Figure 6.

\begin{figure}[t]
\centering
\includegraphics[scale = 0.7]{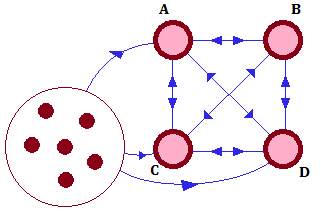}
\caption{Illustration of subgraph segment from which no path to the goal exists, yet contains multiple transition state cycles. Such regions can present non-steady state attractors from which the agent cannot progress to the goal, hence being considered 'trapped' in the subgraph.}
\label{fig:trap_net}
\end{figure}

We can define a subset of states, $tnet$, to represent the states associated with a structure such as this. If we presume to organize $P_g$ such that we align the rows and columns associated with the subnet $tnet$, we can re-cast $P_g$ in the following form, noting that states in $tnet$ can transfer between one another, but not to other states not in $tnet$:
$$
P_g  = 
\begin{pmatrix}
T_{s\notin tnet} & \textbf{0} & \vec{0} \\
T_{s\in tnet} & T_{tnet} & \vec{0} \\
\vec{t}_{g|i \notin tnet} & \vec{0} & 1 \\
\end{pmatrix}
$$

Which we can expand into the successive probability distribution:
\begin{equation}
P^k_g  = 
\begin{pmatrix}
T^k_{s\notin tnet } & \textbf{0} & \vec{0} \\
\Sigma^{k-1}_{j=0} T^j_{tnet}T_{s\in tnet}T^{k-1-j}_{s\notin tnet} & T^k_{tnet} & \vec{0} \\
\Sigma^{k-1}_{j=0}\vec{t}_{g|i \notin tnet}T^j_{s\notin tnet} & \vec{0} & 1 \\
\end{pmatrix}
\label{eqn:tnet_pg}
\end{equation}

From which we can see that $P(s_{i \in tnet} \rightarrow s_g|k)=\vec{0}$ for all k. Further, we will define a system parameter $L_{max}$, the longest minimum length path between any state (from which the goal is reachable) and the goal itself. We have then that for any reachable state $s_i$, $P(s_i \rightarrow s_g|L_{max}) > 0$, or there is a non-zero probability that $s_i \rightarrow s_g$ has occurred after $L_{max}$ timesteps. Consequently we can test if a state is a member of a trap net, as the final row of $P^{L_{max}}_g$ will contain only 0 probability entries in states from which the goal is unreachable. We might use analysis of the connectivity of states in $P_g$ to determine $L_{max}$, however it is sufficient to raise $P_g$ to a power which is greater than $L_{max}$ and examine the final row. In any graph the maximum path length between any two nodes is bounded by the size of the graph, and so it suffices to check $P^{|S|}_g$: any state $i$ for which $P(s_g|\vec{s}_i,k=|S|) = 0$ is necessarily a member of a trap net.

We can then use Equation \ref{eqn:state_prob} to determine the probability at any point in time the system has become stranded in a trap net. Given that the trap states have been identified as above and $P_g$ arranged as in Equation \ref{eqn:tnet_pg}:

$$P(s_t \in tnet|k) = $$
$$
\vec{1}_{1 \times |tnet|}\cdot{}
\begin{pmatrix}
\sum^{k-1}_{j=0} T^j_{tnet}T_{s\in tnet}T^{k-1-j}_{s\notin tnet} & T^k_{tnet} & \vec{0} \\
\end{pmatrix}\cdot{}\vec{s}_t
$$

Which allows for the measure of the risk associated with the system progressing to an inescapable holding pattern, in addition to the probability of achieving the goal state. Additionally, the identification of single attractor states and trap nets together provides a rigorous analysis of the reachability of the goal from all other states as a statistical distribution of time.

\subsubsection{Derivation of bounded time performance}

As a further consideration, we may examine the behavior of the system, as proscribed by the transition matrix, in terms of the evolution of the L1 norm of $T_s^k$. The L1 norm, as all vector induced norms, is submultiplicative, and thus we may write:
$$\|T_s^k\|_1 \leq \|T_s^{k-n}\|_1\|T_s\|_1^{n} $$
$$\|T_s^k\|_1 \leq \|T_s\|_1^k$$

Additionally, because all columns are stochastic, the maximum absolute column sum is paired with the minimum probability single step goal transition. Put otherwise:

$$\|T_s\|_1 = 1 - \min_{\forall a,i} P_g[s_i,g,a]$$
$$\|T_s^k\|_1 \leq (1 - \min_{\forall a,i} P_g[s_i,g,a])^k$$

For many systems, $\min_{\forall a,i} P_g[s_i,g,a] = 0$, which provides little insight. For any reachable state, however, we may once again use $L_{max}$, the maximal shortest path to goal length. Then, at $k=L_{max}$, all states from which the goal is reachable have a  non-zero transition probability, and $L_{max}$ is also the first time step at which all state may possibly have transitioned to the goal state.

\begin{equation}
  \left\{
\begin{array}{ll}
      \|T_s^k\|_1 = 0 & k < L_{max} \\
      \|T_s^k\|_1 \leq \|T_s^{L_{max}}\|_1^{k-L_{max}} & otherwise\\
\end{array} 
\right. 
\label{eqn:thresh_lmax}
\end{equation}

Which allows us to calculate a minimum time at which all states will surpass a certain threshold likelihood of having transitioned to the goal without projecting the system forward in time arbitrarily. Generally for some minimum transition probability threshold $P_{thresh}$:

$$1 - P_{thresh} \leq \|T_s^{L_{max}}\|_1^{k_{p}-L_{max}} $$

\begin{equation}
k_{p} \geq \frac{log(1 - P_{thresh})}{log(\|T_s^{L_{max}}\|_1)} + L_{max}
\label{eqn:kp_limit}
\end{equation}

Which establishes an expectation curve for the progression of states towards the goal in terms of both the dynamics of $T_s$ and the effective 'distance' between the starting state and the goal. Writing the relation slightly differently, as $1 - \|T_s^{L_{max}}\|_1^{k_{p}-L_{max}} \leq  P_{thresh}$, we can also see that the probability of transition to goal grows at an exponential rate, illustrating that even under stochastic disturbances, the path planning algorithm will be highly efficient.

\subsection{Analysis of Abstraction Robustness}

Validation of a learning algorithm against an abstract model is often challenging due to the difficulty of parameterizing abstractions, and the fact that most learning algorithms themselves implement some level of state abstraction inherent to the learning structure. Defining abstraction is a complex topic, and settling on metrics for measuring it even more so. As our system retains all observed information in a probabilistic fashion, it presents a distinct opportunity for evaluation of abstraction as a disturbance model. Towards this end, we will be generalizing the proofs of learning convergence and uncertainty tolerance using a model based on transforms which introduce conflation between states, or reduce the size of the state space.

For our purposes, we will consider an abstracted learning problem to be one in which there is some mapping $\alpha ()$ which transforms a large state space $S$ into a more compact space $\alpha(S)$. $\alpha$ need not be strictly surjective, but for purposes of analysis, we will consider only state pairs in the domain of $S$ and the codomain $\{\alpha(s_i)|\forall s_i\}$, and discrepancies of incompleteness will be modeled in terms of the states which are present in either set. Note that by definition $|\alpha| \leq |S|$.

Presume that we have an $|\alpha| \times |S|$ transformation matrix, $\alpha_{T}$ which contains in each cell $\alpha[j,i]$ the probability $P(\alpha(s_i) = \alpha(j))$ that the $i^{th}$ true state is mapped onto the $j^{th}$ abstracted state. With a given state probability vector $\vec{s}_t$, then, the corresponding probability vector in the abstracted state space is given by $\vec{s}_{\alpha t} = \alpha_T \cdot \vec{s}_t$, or, for general time propagation:
$$\vec{s}_{\alpha t} = \alpha_T \cdot P_g^t \cdot \vec{s}_0 $$

Given a learned AFI subgraph for the abstracted space, $P_{\alpha}$, we also have $\vec{s}_{\alpha t} = P^t_{\alpha}\vec{s}_{\alpha 0}$, and since $\vec{s}_{\alpha 0} = \alpha_T \vec{s}_0$ we can construct a relation from the equivalence $ \alpha_T P^t_g = P^t_{\alpha} \alpha_T$:

\[  \left\{
\begin{array}{l}
        P^t_{\alpha} = \alpha_T P^t_g \alpha^{+}_T \\
        P^t_g  = \alpha^{+}_T P^t_{\alpha} \alpha_T \\
\end{array}
\right. \]

Where $\alpha^{+}_T$ is the pseudoinverse of $\alpha_T$. This pair of transformations allows the conversion from the \textit{probability space} into the probability space for the problem. It is notable that this transform does not allow for conversion into the true \textit{state space}, even if $\alpha_T$ is known perfectly, as $\alpha^{+}_T$ cannot unmix states which are combined, whether stochastically or deterministically. Mathematically, this is realized by $\alpha_{T}^{+}$ not being strictly positive definite.

We can note that:
$$P^t_{\alpha} = \alpha_T P^t_g \alpha^{+}_T = (\alpha_T P_g \alpha^{+}_T)^t$$

for $t=2$, we have that $\alpha_T P^2_g \alpha^{+}_T = \alpha_T P_g \alpha^{+}_T \alpha_T P_g \alpha^{+}_T$, or $\alpha^{+}_T \alpha_T = I$, implying that the columns of $\alpha_T$ must be linearly independent, a natural conclusion given the definition of $\alpha()$ being a surjection on stochastic vectors.

We can thus derive an expanded form for the abstracted transition array in terms of the true transitions by taking the partitions $\alpha_{Ts} = \alpha[:,:-1]$, $\alpha_{Tg}= \alpha[:,-1]$, $\alpha^{+}_{Ts} = \alpha^+[:-1,:]$, and $\alpha^{+}_{Tg} = \alpha^+[-1,:]$, recognizing that both arrays must be stochastic transforms, due to the action on $\vec{s}$:

$$
\alpha_T = 
\begin{pmatrix}
\alpha_{Ts} & \alpha_{Tg}\\
\vec{1}-\vec{1}\alpha_{TS} & 1-\vec{1}\alpha_{Tg}\\
\end{pmatrix}
$$

$$
\alpha^{+}_T = 
\begin{pmatrix}
\alpha^+_{Ts} & \alpha^+_{Tg} \\
\vec{1}-\vec{1}\alpha^+_{TS} & 1-\vec{1}\alpha^+_{Tg} \\
\end{pmatrix}
$$

Transforms between the probability spaces, however, allow us to apply the analysis in Section VI to convergence and learning in the abstracted space. Firstly, we take Equation \ref{eqn:pg_timestep} where we annotate: $\vec{t_{\alpha g}} \cdot{} \sum_{l=1}^{k-1} T^l_{\alpha s} + \vec{t_{\alpha g}} = V_p$:

$$
P^k_g  = 
\begin{pmatrix}
T^k_{s} & \vec{0} \\
\vec{P}(s_i \rightarrow s_g|k) & 1 \\
\end{pmatrix}=
$$
$$
\begin{pmatrix}
\alpha^+_{Ts} & \alpha^+_{Tg} \\
\vec{1}-\vec{1}\alpha^+_{TS} & 1-\vec{1}\alpha^+_{Tg} \\
\end{pmatrix} \cdot{}
\begin{pmatrix}
T^k_{\alpha s} & \vec{0} \\
V_p& 1\\
\end{pmatrix}\cdot{}
\begin{pmatrix}
\alpha_{Ts} & \alpha_{Tg}\\
\vec{1}-\vec{1}\alpha_{TS} & 1-\vec{1}\alpha_{Tg}\\
\end{pmatrix}
$$

Expanding $P^k_g$ lets us calculate the probability of goal transition in the true space:

$$\vec{P}(s_i \rightarrow s_g|k) = $$
$$V_p \alpha_{Ts} - \vec{1}\alpha^+_{Tg}V_p \alpha_{Ts}+ \vec{1}T^k_{\alpha s}\alpha_{Ts} - \vec{1}\alpha^+_{Ts}T^k_{\alpha s}\alpha_{T s}$$
$$+\vec{1} - \vec{1}\alpha^+_{Tg}\vec{1} - \vec{1}\alpha_{Ts}+\vec{1}\alpha^+_{Tg}\vec{1}\alpha_{Ts}$$

using the relations $\vec{1}T^k_{\alpha s} = 1 - V_p$, and $\vec{1}\alpha^+_{Tg} = ||\alpha^+_{Tg}||$ we can re-cast this expression as:

$$\vec{P}(s_i \rightarrow s_g|k) = \vec{1} +||\alpha^+_{Tg}||(\vec{1}\alpha_{Ts}- \vec{1})
$$
$$ - ||\alpha^+_{Tg}||V_p \alpha_{Ts}- \vec{1}\alpha^+_{Ts}T^k_{\alpha s}\alpha_{T s}$$

We presumed that $P_{\alpha}$ is convergent, and thus we can note the limiting behavior of $T^k_{\alpha s}$ and $V_p$:

\[  \left\{
\begin{array}{l}
        \lim_{k \to\infty} V_p = \vec{1} \\
        \lim_{k \to\infty} T^k_{\alpha s} = \textbf{0} \\
\end{array}
\right. \]

From which the limiting behavior of $\vec{P}(s_i \rightarrow s_g|k)$ can be determined:
$$\lim_{k \to\infty} \vec{P}(s_i \rightarrow s_g|k) = \vec{1} +||\alpha^+_{Tg}||(\vec{1}\alpha_{Ts} - \vec{1})- ||\alpha^+_{Tg}||\vec{1} \alpha_{Ts}$$
$$\lim_{k \to\infty} \vec{P}(s_i \rightarrow s_g|k) = \vec{1} - ||\alpha^+_{Tg}||\vec{1}$$

Convergence of $P_g$ can be expressed as $\vec{P}(s_i \rightarrow s_g|k) \rightarrow \vec{1}$, so:

\begin{equation}
\lim_{k \to\infty} \vec{P}(s_i \rightarrow s_g|k) = \vec{1} =\vec{1} - ||\alpha^+_{Tg}||\vec{1}
\label{eqn:ab_conv_cond}
\end{equation}
$$
0 = ||\alpha^+_{Tg}||
$$

Which shows that the convergence of the true system to the goal, given convergence of the abstracted state, is predicated on the transform between the true goal states and the abstracted goal states being \textit{onto}. Note that this does not preclude early convergence due to canceling factors between the dynamics of $V_p$, $T^k_{\alpha s}$, and $\alpha^+_{Tg}$, but rather determines the asymptotic behavior of the system.

\subsection{Impact of Abstracted State on Performance}

Given this condition on the abstraction function, and presuming again that the abstracted $P_{\alpha}$ is convergent, we can further extend the derivations in Section 4 to the case of learning on an abstracted system.

Beginning with the relation $||T^k_s||_1 \leq ||T_s||^k_1$ for the true state system:
$$
||T_s||^k_1 \geq
$$
$$
||\alpha^+_{Ts}T_{\alpha k}\alpha_{Ts}+\alpha^+_{Tg}V_p\alpha_{Ts}+\alpha^+_{Tg}\vec{1}-\alpha^+_{Tg}\vec{1}\alpha_{Ts}||_1
$$
$$
\geq
||\alpha^+_{Ts}T_{\alpha k}\alpha_{Ts}||_1
+||\alpha^+_{Tg}||_1(1-||T_{\alpha k}||_1||\alpha_{Ts}||_1)
$$

Because $\alpha^+_{Tg}$ is a vector, $||\alpha^+_{Tg}||_1 \geq ||\alpha^+_{Tg}||$, and $||T_{\alpha k}||_1$, $||\alpha_{Ts}||_1$ are submatricies of stochatic matricies, they are strictly in $[0,1]$ (though this is not the case for $||\alpha^+_{Ts}||_1$, and so this proof applies only $P_{\alpha}\rightarrow P_{g}$ and not to $P_g\rightarrow P_{\alpha}$) (that is, convergence of the abstracted model implies convergence of the true model, but not the converse), so:

$$
||T_s||^k_1 \geq
$$
$$
||\alpha^+_{Ts}T_{\alpha k}\alpha_{Ts}||_1
+||\alpha^+_{Tg}||(1-||T_{\alpha k}||_1||\alpha_{Ts}||_1) \geq
$$
$$
||\alpha^+_{Ts}T_{\alpha k}\alpha_{Ts}||_1 = ||\alpha^+_{Ts}||_1\cdot{}||T_{\alpha k}||_1\cdot{}||\alpha_{Ts}||_1
$$

From this inequality, we can then replicate the prior analysis for the abstracted case:

$$\|T_s^k\|_1 \leq (1 - \min_{\forall a,i} P_g[s_i,g,a])^k$$
$$1 - P_{thresh} \leq (||\alpha^+_{Ts}||_1\cdot{}||T_{\alpha k}||_1\cdot{}||\alpha_{Ts}||_1)^{k_{p\alpha}-L_{max}} $$
\begin{equation}
k_{p\alpha} \geq \frac{log(1 - P_{thresh})}{log(||\alpha^+_{Ts}||_1\cdot{}||T_{\alpha k}||_1\cdot{}||\alpha_{Ts}||_1)} + L_{max}
\label{eqn:kp_lim_ab}
\end{equation}


Which describes how the inclusion of the abstraction modifies the minimum expected time to achieving the goal state relative to the timescale predicted by $P_{\alpha}$ alone, or put alternatively, presuming the learned state space fully represents the behavior of the system without hidden variables. 

By examining the expression above, we can make some inferences about the impact of $\alpha_{T}$ on convergence performance:

\begin{equation}
  \left\{
\begin{array}{ll}
        k_{p\alpha} > k_p & ||\alpha_{Ts}||_1\cdot{}||\alpha^+_{Ts}||_1 < 1 \\
        k_{p\alpha} \leq k_p & ||\alpha_{Ts}||_1\cdot{}||\alpha^+_{Ts}||_1 \geq 1\\
\end{array}
\right. 
\label{eqn:kp_shift}
\end{equation}

We can use the product above as a rough measure of the 'quality' of an abstraction,  the degree to which it effects performance, by:

$$Q(\alpha_{T}) = \frac{1}{||\alpha_{Ts}||_1\cdot{}||\alpha^+_{Ts}||_1}$$

So that $Q(\alpha_T)$ is directly correlated to the impact $\alpha_T$ has on performance. It is worth mentioning that because $\alpha^+_T$ is not strictly positive definite, conditions under which $Q(\alpha_{T})$ improve system performance are possible, albeit difficult to design. This suggests that the introduction of an abstraction may either improve or reduce efficacy, depending on the nature of the abstraction. Improvements may seem counter-intuitive, but consider the way a substitution may reduce the number of steps needed to solve an algebraic equation. In the context of planning, certain simplifications may indeed bias portions of the graph towards choosing probabilistically identical, but shorter paths, minimizing the outlier chances of stochastic variance increasing average path-to-goal length.

Empirically, we can also approximate this measure by calculating the constituent components of the relation between $k_p$ as predicted for the system under abstraction and the measured $k_{p\alpha}$ as the average number of steps to reach the goal over many iterations:
$$
\frac{k_{p\alpha}-k_p}{k_p-L_{max}}= 
\frac{log(||\alpha^+_{Ts}||_1\cdot{}||\alpha_{Ts}||_1)}{log(||T_{\alpha k}||_1)}
$$
\begin{equation}
||T_{\alpha k}||_1^{\frac{k_p - k_{p\alpha}}{k_p-L_{max}}}= 
Q(\alpha_T)
\label{eqn:Q_emp}
\end{equation}

Which allows us to calculate a measure of the relative efficacy of the abstraction from the learned transition array, maximum shortest path, and measured minimum and average path lengths across samples. Further, in Section 5, this relation will allow us to examine the expected learning curves for the agent under training.

\subsection{Learning Convergence}

We can evaluate the behavior of the agent as a learning system by modeling the learned AFI matrix as an abstracted function function of the true state. We may begin by presuming a transform which maps the true states onto a distribution which reflects the initial assumptions of the learning model-- namely, a uniform distribution from which actions are initially chosen randomly:

\[  \left\{
\begin{array}{l}
        \alpha_{T1} = \frac{1}{|\alpha|}\cdot{}\textbf{1} \\
        \alpha^+_{T1} = \frac{1}{|S|}\cdot{}\textbf{1} \\
\end{array}
\right. \]

With these, we can see that:
$$
\begin{array}{ll}
        ||\alpha_{T1}||_1 = \frac{|\alpha|-1}{|\alpha|} & ||\alpha^+_{T1}||_1 = \frac{|S|-1}{|S|} \\
\end{array}
$$

Further, presuming that $P_{g}$ is the asymptotic learning goal matrix and $P_{\alpha}$ is the non-learned array, we have that $|\alpha| = |S|$.

We can approximate the expected learning curves over many samplings by presuming, from random choice prevailing at non-sampled occasions, an amortized update at each step $k$ derived from Equation \ref{eqn:apost_prob}. Examining an update to a single state vector, we can note that in the average case at step $k$ the state has been visited $\frac{k}{|S|}$ times. Given the probability vector $\vec{s}_{\alpha i}$ from $P_{\alpha}$, then, the individual counts can be expressed as $\frac{k}{|S|}\vec{s}_{\alpha i}$, also via Equation \ref{eqn:apost_prob}. Further, the probability distribution for the increase in counts can be expressed by $\vec{s}_i$ (the asymptotic learned behavior), the corresponding expectation of the column in $P_g$. Combining the prior recorded occasions with the new, for $\frac{k+1}{|S|}$ steps gives:

$$
\vec{s'}_{\alpha i} =  \left( \frac{k}{|S|}\vec{s}_{\alpha i} + \frac{1}{|S|}\vec{s}_i \right)\cdot{}\frac{1}{\frac{k}{|S|} + \frac{1}{|S|}} = 
\frac{k\vec{s}_{\alpha i} + \vec{s}_i}{k + 1}
$$

$$
\delta \vec{s}_{\alpha i} = \vec{s'}_{\alpha i} - \vec{s}_{\alpha i} = 
\frac{\vec{s}_i - \vec{s}_{\alpha i}}{k + 1}
$$

Which, in aggregate, gives the expression across the full transition array:

$$
\delta P_{\alpha k} = \frac{P_g - P_{\alpha k}}{k + 1}
$$

For the recurrence relation:

\[  \left\{
\begin{array}{l}
        P_{\alpha k+1} = \frac{kP_{\alpha k} + P_g}{k + 1} \\
        P_{\alpha 1} = \frac{1}{|S|^2}\cdot{}\textbf{1}\cdot{}P_g\cdot{}\textbf{1} \\
\end{array}
\right. \]

\begin{equation}
P_{\alpha k} =
\left[\frac{\textbf{1}\cdot{}P_g\cdot{}\textbf{1}}{k|S|^2} + \frac{k-1}{k}P_g\right] =
\alpha_{Tk} P_g \alpha^+_{Tk}
\label{eqn:pak_rel}
\end{equation}

Which we can express in similar block fashion as above:
$$
\frac{1}{k|S|}
\begin{pmatrix}
\textbf{1} + |S|(k-1)T_s & \vec{1} \\
\vec{1}+|S|(k-1)P(g) & 1+|S|(k-1) \\
\end{pmatrix}
\alpha_{Tk}
= 
\alpha_{Tk}P_g
$$

And calculate the non-goal block of each side, using the general form for $\alpha_{Tk}$
$$
\left(\frac{1-k|S|}{k|S|}\textbf{1} + \frac{k-1}{k}T_s\right)
+ \textbf{1}\alpha^+_{Ts}
=
\alpha_{Ts}T_{\alpha s}\alpha^+_{Ts} + \alpha_{Tg}V_P\alpha^+_{Ts}
$$

$$
\left[\frac{\textbf{1}_{|S|-1}}{k|S|} + \frac{k-1}{k}T_s\right] 
-\alpha_{Tg}\vec{1}+\alpha_{Tg}\vec{1}\alpha^+_{Ts}
$$
$$=
\alpha_{Ts}T_s\alpha^+_{Ts}+\alpha_{Tg}P(g)\alpha^+_{Ts}
$$

In the limit case, $k \rightarrow \infty$, the right side of both above are equivalent, which allows us to equate and simplify:

$$
(\textbf{1} - \alpha_{Tg}\vec{1})\alpha^+_{Ts}
=
\textbf{1}
-\alpha_{Tg}\vec{1}
$$
$$
\alpha^+_{Ts} = \textbf{I} \rightarrow \alpha_{Ts} = \textbf{I}
$$

Demonstrating conclusively that as $P_{\alpha}$ is learned, the corresponding abstraction transforms approach the identity matrix, and thus learning converges on basis of Equations \ref{eqn:apost_prob} and \ref{eqn:mag_conv}.

Equation \ref{eqn:pak_rel} additionally provides the expression of form for the learning curve associated with the GAP learning phase. We can express it as:

$$
P_{\alpha k} = \frac{1}{k}
\left(
\frac{\textbf{1}}{|S|} - P_g
\right)
+ P_g
$$

In which the terms $\frac{\textbf{1}}{|S|}-P_g$ and $P_g$ are clearly time invariant, with the dynamic behavior explicitly governed by the reciprocal of the timestep, $\frac{1}{k}$, and thus the average learning curve will follow a reciprocal pattern $k_{p\alpha}(k) = A\frac{1}{k} + B$. $B$ is naturally the asymptotic average path to goal length, $k_p$. To determine $A$, we can evaluate the initial behavior of the system given the form for $\alpha_{T1}$ and Equation \ref{eqn:pak_rel}:
$$
k_{p\alpha}(1) - k_p = A
$$
$$
A= (k_p-L_{max})\frac{2log(|S|)-2log(|S|-1)}{log(||T_{\alpha 1}||_1)}
$$

$||T_{\alpha 1}||_1$ can be directly calculated from $\alpha_{T1}P_g\alpha^+_{T1}$ as $\frac{|S|-1}{|S|}$, and thus:
$$
A= 2(L_{max} - k_p)
$$
\begin{equation}
k_{p\alpha}(k) = \frac{2(L_{max} - k_p)}{k} + k_p
\label{eqn:lin_learn}
\end{equation}

Which establishes an expected average for the performance of the unlearned system as a biased random walk on $P_g$, with $k_{p\alpha}(1) = 2L_{max} - k_p$, as well as the general form for the learning curve of the GAP algorithm being an offset reciprocal function of step number.

\section{Demonstration Cases}

In this section we explore the application case behavior of GAP based agents learning in three example problem classes: a classical STRIPS type problem, a complex joint domain of the TAXI domain and Maze Navigation problems, and the Tower of Hanoi puzzle. These tasks includes specific hierarchical components, complex and large state spaces, and have been used previously as benchmark trials for machine learning algorithms. For instance, the TAXI domain by \cite{dietterich1999state}, Mazes by \cite{mccallum1995reinforcement}, and the Tower of Hanoi by \cite{knoblock1990abstracting}.

Using these cases, we seek both to illustrate the efficacy and efficiency of the GAP algorithm across multiple problem domains, validate the analytic models illustrated in Section 4, and evaluate relations between the predictions made therein.

\subsection{Experimental Procedures}

Prior to detailing the experiments themselves, we outline the common procedures used across all trials which are not specific to any one problem case.

\subsubsection{Training Process}

To train the agent, the AFI datastructure is initialized with a uniform distribution of random values, as expressed analytically in Section 4. Upon observations of occasions, the corresponding INC cells are updated, with the AFI array modified proportionally to the actual number of measured values, as per Equations \ref{eqn:apri_prob} and \ref{eqn:apost_prob}. The agent proceeds in the simulation environment acting on basis of the current state of the AFI array until achieving the goal state, which terminates the epoch, upon which event the simulation world is reset, with the INC array persisting between epochs.

\subsubsection{Error Induction}

To investigate system performance under uncertainty, we also artificially induce error at certain points. Error induction is achieved in planned execution by a random threshold process which, a select proportion of time, executes a hidden non-planned action in lieu of that in the plan. Action modifications are selected as the medium for two reasons: first, discontinuous transitions within the world do not model real world uncertainties well. For instance, a mobile robot may accidentally slide, changing position as though it had selected a different trajectory, but it is unlikely to teleport. Second, because modeling of a state estimation error can be approximated well in most cases by an action change: if there is an error in state detection, then the agent will instead take action based on the fault state, which under a uniform distribution error model will be an action independent of the current state.

\subsubsection{State Generation}

Because we are intending to make a fully abstracted learning agent, every problem is built without labeled states, simply initialized to a rough estimate number of possible states. The simulation models are designed to output string type state reports when polled for information, and from these strings, a simple hash algorithm generates a lookup table for the agent to use. As more states are discovered, the algorithm iterative increases the size of the SSA arrays and corresponding array slice structures to accommodate larger lookup tables. 

\subsubsection{Calculation of Metric parameters}

Throughout this section, we will be demonstrating the effectiveness of the GAP algorithm by measuring a suite of parameters related to performance characteristics. First and foremost of these are the metrics which represent the measured learning curves as a reciprocal functions as predicted in Section 4, Equation \ref{eqn:lin_learn}. Towards this end, we calculate best fit equations and measure of their accuracy: $R^2$ for the fit of $k_{p\alpha} = \frac{A}{k}+B$, and percentage off linear averages of linear regressions on the plots of $(\frac{1}{k},k_p)$ calculated for $N$ sequential data points on the curve as:

$$\sum_{\forall n \in N} \frac{1}{N}\frac{|k_p[n] - (A\frac{1}{n} + B)|}{k_p[n]}$$

representing the average proportional deviation from the linear approximation for the best fit line.

In addition to these measures to validate the predicted learning curve, we also calculate and compare approximations of $k_p$ and $L_{max}$ to establish a correlation between the analytic predictions in Section 4 and the measured data. 

For $k_p$, calculations are made both by computation of the fit curve for Equation \ref{eqn:lin_learn}, and by averaging performance levels after system convergence. $L_{max}$ comparisons are made between two predicted relationships, Equation \ref{eqn:Q_emp} directly, and Equation \ref{eqn:kp_limit} by estimation over multiple probability levels of induced error for $P_{thresh}$. As $L_{max}$ is a problem constant, the inequality Equation \ref{eqn:kp_limit} can be used to empirically identify the values for $P_{thresh}$ at which the inequality is no longer valid, and thus infer an estimate of $L_{max}$ based on measurements of  $||T_{s}||$.

\subsection{STRIPS-type Problems}

\begin{figure}[t]
\centering
\includegraphics[scale = 0.6]{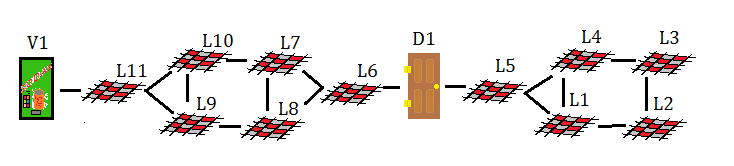}
\caption{Illustration of a basic STRIPS-style world, containing linked location states ($L_i$) and multiple independent actionable states ($V_1$ and $D_1$)}
\label{fig:STRIPS_exp}
\end{figure}

In this section, we illustrate both the learning performance and the planning action on a collection of discrete problem cases built on the hierarchical classes on which the STRIPs algorithm and its successors operate. With these cases, we demonstrate core properties of the algorithm-- efficient learning fitting the predicted reciprocal form and correlation between the learned system and the expected asymptotic behavior.

\begin{figure}[b]
\centering
\includegraphics[scale = 0.5]{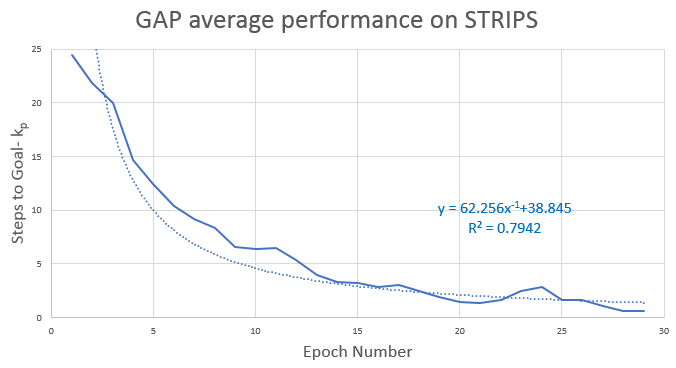}
\caption{Average learning curve over instances of varying error from 0\% to 50\% in the STRIPS problem space, with the reciprocal fit curve plotted superimposed}
\label{fig:STRIPS_learning_basic}
\end{figure}

We begin by implementing a typical STRIPS-type planning problem, as schematically represented in \autoref{fig:STRIPS_exp}. Here, the agent is in a world with a specific set of move operations which translate it through a location space, a pair of world manipulating actions (to fetch an item and open a door), and a state space reflecting an internal state (possession of an item), an external state (location) and a hidden state (status of the door). This represents a hierarchically ordered workspace, critical because expressing functionality within such problems is a hurdle for all learning algorithms.

Displayed in \autoref{fig:STRIPS_learning_basic} is the average learning curve, derived over 1000 iterations beginning from no training, and across random induced error levels from 0\% to 50\%. This curve is for an online implementation, with random starting locations in the right half of the world. On this plot we see a reciprocal fit curve of $k_p(k) = 62.3\frac{1}{k} + 38.9$ at $R^2 = 0.79$, and asymptotic learned performance being approximately 39 steps between the starting state and the goal, compared to the no error absolute minima of 17.

\begin{figure}[t]
\centering
\includegraphics[scale = 0.5]{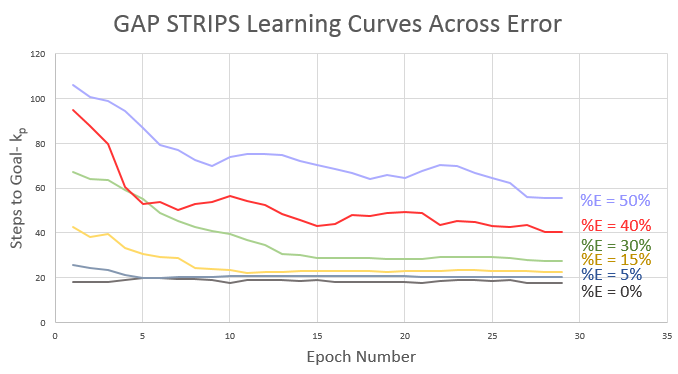}
\caption{Learning curves for the STRIPS problem across levels of induced error from 0\% to 50\%}
\label{fig:STRIPS_learning_err}
\end{figure}

\begin{figure}[t]
\centering
\includegraphics[scale = 0.55]{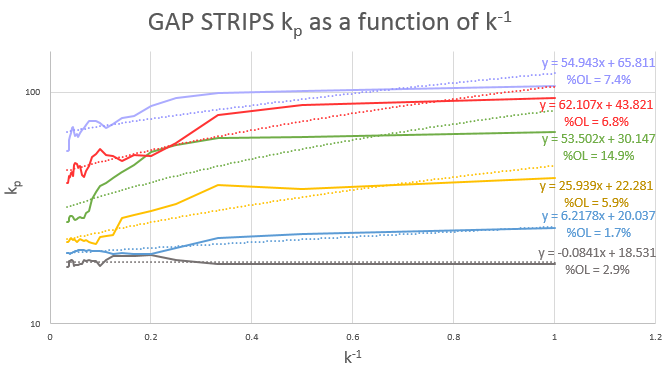}
\caption{Plots of $k_p$ as a function of $\frac{1}{k}$ for various error levels, along with measures of the deviation from linearity in terms of \% off Linear behavior, showing close correspondence to the predicted learning curve form of $k_p = A\frac{1}{k}+B$}
\label{fig:STRIPS_recip_kp}
\end{figure}

Figure \ref{fig:STRIPS_learning_err} showcases the learning curves of the GAP algorithm on this problem at each induced error level independently. Each curve is the average performance at each epoch as averaged over 50 trials. We can see from these curves that the learning tends to follow the same reciprocal pattern as the general curve, with variance in asymptotic performance shifting due to the increase in error rate elevating the expected number of steps to reach the goal. To reinforce the reciprocal relationship, we also plot the linearization of these curves (in Figure \ref{fig:STRIPS_recip_kp})  the number of steps to reach the goal plotted against $\frac{1}{k}$, along with the off linear percent labeled for these curves. For each plot but one, the deviation from linear fit is in the single digits, with the greatest deviation being for the 30\% curve, with a 15\% average off linear error. These measurements serve to validate the prediction of Equation \ref{eqn:lin_learn} that the GAP algorithm will express reciprocal learning curves.

In addition to these linearized plots showing correlation between $\frac{1}{k}$ and steps to goal, we also highlight the correlations between $k_p$ as predicted by the asymptotic behavior of the data itself and the fit reciprocal curve. We also calculate $L_{max}$ from Equation \ref{eqn:lin_learn} and as predicted by the threshold in Equation \ref{eqn:thresh_lmax}. Both measures are presented on Table 2, along with the corresponding percent errors. Here, we can see that the differences between the asymptotic $k_p$ and the fit function are small, ranging from 0.87\% to 7.12\% for induced errors up to 40\%, and the difference between the the measured and predicted $L_{max}$ is 7.8\%, indicating very close correspondence between the observed performance and the predictions of Equations \ref{eqn:lin_learn} and \ref{eqn:thresh_lmax}.

\begin{table}[b]{
\centering
\begin{tabular}{|c|c|c|c||c|c|}
\hline
$P_{thresh}$  & $k_p$ Meas: & $k_p$ Pred: & \%E & &$L_{max}$\\
\hline
0\%     & 18.53  & 18.10 & 2.30\%  & Meas: & 25.30\\
\hline
5\%     & 20.04  & 20.21 & 0.87\%  & Pred: & 27.29\\
\hline
15\%     & 22.81  & 22.76 & 2.11\%  & \%E & 7.8\%\\
\hline
30\%     & 30.15  & 28.14 &  7.12\% &  & \\
\hline
40\%     & 43.82  & 42.07 & 4.15\%  &  & \\
\hline
50\%     & 65.81  & 57.40 & 14.65\%  &  & \\
\hline
\end{tabular}
\caption{Comparison of measured and predicted values for systematic form analysis as in Section 4, as calculated from the performance on the STRIPS problem learning curves}
}
\label{tbl:STRIPS_preds}
\end{table}

What the successive curves illustrate at the higher levels of induced error is that there is a close hew to Equation \ref{eqn:lin_learn}, even as the learning process is disturbed by a persistent error signal which distorts the true model-- essentially an abstraction with a substantially small $Q(\alpha_{T})$.

Of note is the 50\% error case, for which the discrepancy is larger, roughly twice the level of the next greatest deviation. However, an introduction of 50\% error into the action of the agent is extremely substantial, and it is reasonable to expect that the learning performance will degrade. Qualitatively speaking, as the induced error rate increases, $P_{\alpha}$ behaves more and more like a random uniform stochastic process than the underlying 'true system' $P_g$. Referring back to Equation \ref{eqn:Q_emp}, we can see that for a certain magnitude of error (expressed in terms of $||\alpha^+_{Ts}||_1$), the limit of $k_{p\alpha}(k)$ will grow to the point at which the difference between $k_{p \alpha}(0)$ and the asymptotic performance is effectively negligible. Because of the exponential form of Equation \ref{eqn:Q_emp}, the level will be highly sensitive to the exact value of $||\alpha^+_{Ts}||_1$, but it represents a threshold at which learning is no longer effectual. In more rigorous terms, $\lim_{k \rightarrow \infty} k_{p\alpha}(k) \rightarrow L_{max}$, and so the function $k_{p \alpha}(k)$ no longer properly behaves as a reciprocal, but as a constant function, exactly the expected behavior of an attempt to learn a uniform random process.

In the late stages of learning for each of the error variant curves the performance levels are clearly tiered in proportion to the error rate, which shows that increasing uncertainty directly increases the average number of steps needed to reach the goal state, as predicted by Equation \ref{eqn:Q_emp}, while preserving the learning curve form. Further, we have the precise convergent parameters, $k_p$, illustrated in Table 2, both as calculated by average of terminal cases, and as the asymptotic of the fit curves, showing the consistent increase as a function of $P_{thresh}$.

\begin{figure}[t]
\centering
\includegraphics[scale = 0.60]{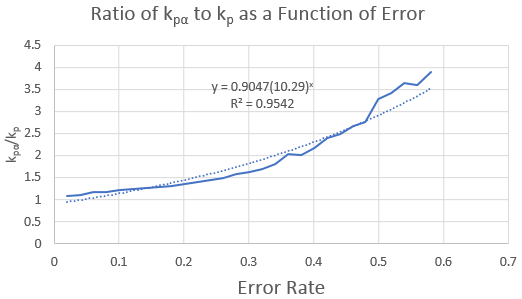}
\caption{Plot of $\frac{k_{p\alpha}}{k_p}$ ratio for the learning curves of the GAP algorithm on the STRIPS world, along with the power law fit curve}
\label{fig:STRIPS_kp_ratio}
\end{figure}

We can use this data to confirm the predicted relationship suggested in Equations \ref{eqn:kp_limit} and \ref{eqn:Q_emp}. In particular, if we take $k_p$ to be the convergent performance of the 0\% error case (considered reasonable on account of the problem simplicity and closeness of the measured value of 18 steps to the theoretical minimum case of 17 steps) we can examine the ratio of $k_{p\alpha}$ at multiple error levels to this baseline $k_P$. Plotting the ratio $\frac{k_{p \alpha}}{k_p}$ in Figure \ref{fig:STRIPS_kp_ratio}, the proportional change between the asymptotic performance of the induced error case and the error free case. When we examine this ratio, we find that there is a high correlation exponential relationship between the error rate and the terminal performance of the agent, mirroring the predicted relationship.

In Table 2, we also show $L_{max}$ as derived via both methods described in Section 5.1.4, and the corresponding error between the two. Both of these sets of measures present a close correspondence between the analytical predictions derived in Section 4 and the estimated values of the simulated problem. Further analysis of the impact of this measure is limited as the values are identified by successive increases in $P_{thresh}$ to identify the limit for Equation \ref{eqn:kp_limit} and hence we can only identify one data point across the problem space. However, the closeness of the two values serves, with the attendant prior confirmations, to validate the analysis.

The key observations across this entire set of results is that the GAP agent consistently learns the problem structure, even across high degrees of artificially introduced error to the problem. Additionally, the learning curves closely follow the predicted forms in Section 4, and derived measures from the analytics in that section are also in close adherence to those measured within the problem space, validating these predictions for the STRIPS problem, and enabling us to investigate further on more complex problems in the next section with the knowledge that the learning performance is as expected.

\subsection{Maze/TAXI Domain}


The TAXI and Maze problems are canonical study cases for machine learning systems. In the TAXI problem, the agent must visit a list of locations, pick up a 'passenger', and then deliver each passenger to a specific destination cell. We additionally complicate the problem by performing the navigation component inside a maze. These mazes can be well conditioned mazes, such as those lacking open fields which confound wall following algorithms, and ill conditioned ones which possess such areas. Combining the two problems creates a complex hierarchical problem of similar character to the STRIPS implementation, but with substantially larger state spaces and far more complex learning patterns, concurrent with ample opportunity for error induction and abstraction onto the learning case.

For an agent, the canonical operations are movements in each of the four cardinal directions, and pickup and drop off actions which can only be performed at explicit locations within the maze. Inputs to the system, naturally, vary depending on the abstraction mechanisms being employed, but in general include local observations of the maze topography in some region near to the agent, registration of the relative position of the target 'passenger', and whether a passenger is currently carried.

Of particular note for our experiments here (speaking to the goal agnosticism of the GAP algorithm)is that we do not perform training on fixed TAXI destinations and mazes, but rather generate a random maze for each training epoch, complete with random target locations for 'pickup' and 'drop off' actions. The inclusion of both TAXI elements and Maze elements allows us to better explore the functioning of the GAP algorithm in complex hierarchical problem spaces. It is possible to see that, taxonomically, a combination of navigation and sequence ordering elements such as this is, essentially, an expanded case of the STRIPS type problem. By increasing the complexity, we are able to examine higher order behavior as an extension of the observations made in the prior experiments.

\subsubsection{Simple Maze/TAXI Domain}

\begin{figure}[b]
\centering
\includegraphics[scale = 0.50]{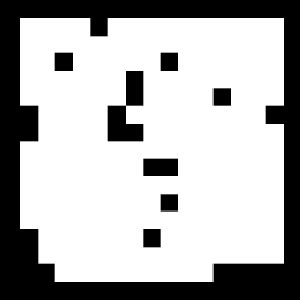}
\caption{Example TAXI Domain world of the type used in these experiments, with sparse obstacles placed (covering ~10\% of the working area) and with a regional spacing of 15x15 cells}
\label{fig:MAZE_TAXI_error_curves}
\end{figure}

\begin{figure}[t]
\centering
\includegraphics[scale = 0.55]{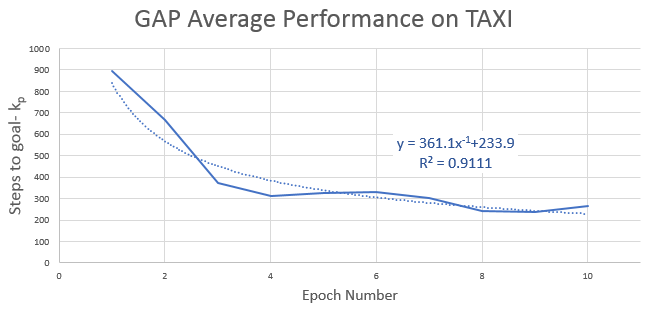}
\caption{Average performance of the GAP algorithm learning the simple implementation of the Maze/TAXI joint domain problem on a 15x15 'maze' with three randomly placed target pickups and drop offs}
\label{fig:avg_taxi_learn}
\end{figure}

To begin with, we examine performance in a relatively simple field of activity, where the 'maze' component is more akin to obstacles, as illustrated in Figure \ref{fig:avg_taxi_learn}. These are areas of 15x15 cells with some number of randomly placed 'wall' cells, and three target pickup and dropoff locations. Agent actions include cardinal direction movements, pickup and dropoff actions, and the states are constructed from the location of the agent, whether the agent currently has a passenger, whether the agent is at a passenger target, destination, or neither, and the number of remaining passengers.

This simple domain allows us first to validate that the agent will successfully learn in this sort of hierarchical with substantially more states, paving the way for the study of the more complex forms of the problem. It also presents a chance to investigate the impacts of input abstraction on the learning in the absence of artificial error. State abstractions would be possible in the STRIPS-type problem, however the state space is so small that proportional changes to $|\alpha_{Tg}|$ are, in practice, difficult to compare numerically.

Towards that end, we explore three additional state construction mechanisms: reduction of both location coordinates by a factor of 2 (essentially reducing the navigation to 2x2 blocks in the greater world). Reduction of just the horizontal position coordinate by a factor of 3, and reduction of both coordinates by $1\frac{1}{2}$. Each of these abstractions reduces the state space size by compressing the navigation space, at the cost of mixing cells which are mapped together into the same probability state.

We first plot the learning curve over the aggregate of all experiments (numbering 50 per abstraction, for 200 total trials) in Figure \ref{fig:MAZE_TAXI_error_curves}, to reinforce that the aggregate performance of the agent in this world matches the predicted learning behavior. As before, we see a reciprocal fit with a high correlation of $R^2 = 0.91$, better matched than that of the STRIPS case. The asymptotic ideal performance case is 234 steps across all trials, with the measured peak no abstraction performance being 222 steps. This observation alone is interesting, as it suggests that the GAP algorithm is, among these abstractions, able to bring the performance curves for the abstracted state spaces to within about 5\% of the asymptotic  non-abstracted performance level. This implies that the $Q(\alpha_{T})$ factor for these abstractions is substantially better than for the pure random errors; a qualitative argument for the efficacy of the abstractions themselves, and thus their utility as tools for examining system behavior distinct from induced error.

\begin{figure}[b]
\centering
\includegraphics[scale = 0.55]{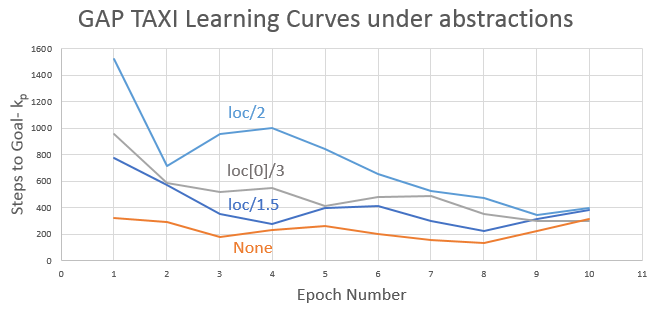}
\caption{GAP learning curves using location based state space compressing abstractions}
\label{fig:simple_maze_abst}
\end{figure}

Figure \ref{fig:simple_maze_abst} presents the learning curves acting on each of the abstractions independently. Though the asymptotic performance limits are comparable, the long term behavior of the curves themselves are substantially varied, with a direct relationship between the effectiveness of the abstractions in learning performance; the 'loc/2' abstraction having the most negative impact, followed in order by 'loc[0]/3', and then 'loc/1.5', with the effective number of initial steps to the goal at the first epoch being directly correlated to learning performance throughout the trials, a trend observable in the STRIPS-style problem, and one which we will continue to see in latter experiments.

\begin{table}[b]{
\centering
\begin{tabular}{|c|c|c|c|}
\hline
Abst.  & $k_p$ Meas: & $k_p$ Pred: & \%E \\
\hline
loc/1.5     & 307.1 & 245.1 & 20.2\% \\
\hline
loc/2       & 405.5 & 421.5 & 3.9\% \\
\hline
loc[0]/3    & 317.1 & 304.5 & 3.9\% \\
\hline
None        & 222.7 & 192.9 & 13.4\% \\
\hline
\end{tabular}
\caption{Measured and model predicted parameter comparison for the GAP algorithm learning the Maze/TAXI Domain problem}
}\end{table}

Table 3 presents the measured and predicted $k_p$ for this problem with each of the state abstraction cases, along with the corresponding errors. In this data, we find a curious relationship between the scale of the error and the abstractions: the no abstraction case and 'loc/1.5' case, which have the best pair of asymptotic learning performance levels, also have the highest proportional discrepancy. On the one hand, the superior performance means that the impact of differences is magnified, however the scale of the errors, 13\% and 20\%, is unlikely to be accounted for by this alone. 

\begin{figure}[t]
\centering
\includegraphics[scale = 0.55]{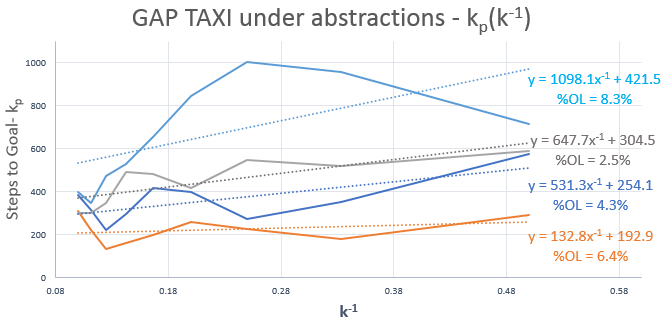}
\caption{Plot of $k_p$ as a function of $\frac{1}{k}$, along with concurrent linear fit as Eq. (*) and percentage of off linear components}
\label{fig:simp_taxi_kp_recip}
\end{figure}

We may, however, gain some additional insight by exploring the fit curves themselves, plotted on Figure \ref{fig:simp_taxi_kp_recip}. Here, we can see that the off linear errors for each curve are relatively low, all less than 10\%, indicating solid close to form adherence to the model of Equation \ref{eqn:lin_learn}. However, the errors associated with the no abstraction and 'loc/1.5' abstractions are approximately twice that of the other pair. Given that these two are the higher performing instances, we may thus hypothesize that the error discrepancy is likely due to fit errors associated with the agent reaching asymptotic performance levels earlier than the slower learning cases. When we make this assumption, and truncate the cure fit at 5 epochs, rather than 10, we find $k_p$ of 288.7 and 205.5 for the 'loc/1.5' and no abstraction cases, respectively, with corresponding errors of 5.9\% and 7.7\%, comparable to the performance of the other pair of trials.

These results show that, as with error induction, it is possible for the algorithm to learn in a reduced state space as generated by abstraction mechanisms. As with the observation in the prior section, drastic reductions in the representativeness of the abstraction may lead to cases in which learning efforts are trivial, but by examining these relatively conservative models for state space reductions, we have been able to illustrate that learning under abstraction is effective in practice.

\subsubsection{Complex Maze/TAXI Domain}

\begin{figure}[b]
\centering
\includegraphics[scale = 0.50]{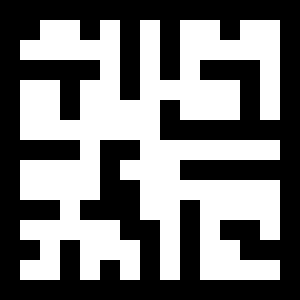}
\caption{An example of a randomly generated ill conditioned maze used in these Maze/TAXI problems}
\label{fig:MAZE_example}
\end{figure}

Building on the validation of the capability of learning under abstracted state spaces in the prior subsection, we now turn to investigating the substantially more complex working space of a full maze. Such a maze is illustrated in Figure \ref{fig:MAZE_example}, which highlights a few particular complications we have produced: rather than restricting ourselves to simple mazes without interior spaces, we have allowed fort the inclusion of open space regions in the maze. Those mazes with uniform width traversals present a set of inherent relationships which make them amenable to solution by simple form maze navigation algorithms. 

This indicates that there is an inherent state space simplification embedded in them, and thus we remove this constraint from the maze generation algorithm to deliberately increase problem complexity, and therefor the sensitivity of our experiments to impacts of varying error and abstraction. 

\begin{figure}[t]
\centering
\includegraphics[scale = 0.5]{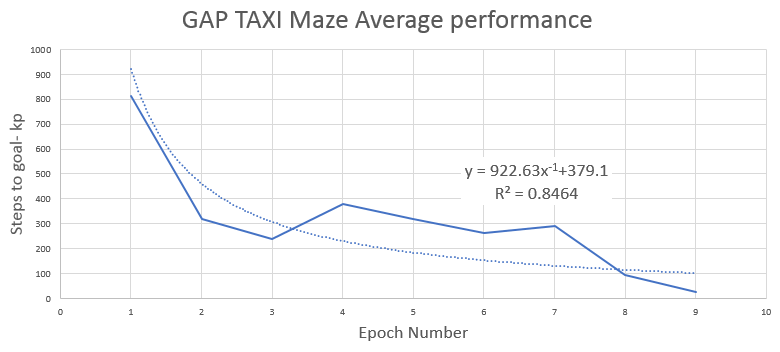}
\caption{Average learning curve for the GAP algorithm operating on the Maze/TAXI problem}
\label{fig:MAZE_TAXI_avg}
\end{figure}

We also elect to represent the state vector as a relative measure. In the most fundamental case, we implement the state model based on the local environment phrased as available movements, a relative vector towards the next objective, and whether or not the agent currently has a 'passenger'. In implementing relativistic states such as this, it becomes  possible to learn the problem in a more general sense than a specific sequence of fixed tasks. This generalized formulation then allows us to examine properties of learning transference and generalization, especially valuable because as in the prior case, each \textit{epoch} is trained in a different maze, with new, random objective locations.

Figure \ref{fig:MAZE_TAXI_avg} shows the average learning curves for this form of the Maze/TAXI problem across trials with error ranging from 0\% to 30\%. With an $R^2$ of 0.84, less than the previous Maze/TAXI case, but still greater than the STRIPS trials. A primary contributor to the error in fit quality is the presence of some outlier learning cases within the data, with the scale of these rare disturbances visible on Figure \ref{fig:MAZE_TAXI_avg} as the sharp jump from epochs 4 to 7. This can be readily observed to be due to the randomization of the maze, as occasionally the random maze presents radically different structures to an agent instance which had previously learned on fairly similar mazes, and must adapt to an expanded problem space. The overall trend towards the asymptote, however, is clear evidence that in the face of this adaptation challenge the GAP algorithm is able to learn the more expansive problem after a few epochs. Further, of particular note is that the amortized scale of the disturbance is much lower than that of the initial performance prior to any learning. This indicates that a measure of learning transference is necessarily occurring, a phenomena we will examine in more detail later.

In Figure \ref{fig:MAZE_TAXI_error_curves}, we plot the curves for the GAP algorithm learning the Maze/TAXI problem across levels of induced error ranging from 5\% to 30\%. Present here are two previously noted trends: the relationship between the asymptotic $k_p$'s proportionality to the error rate, and the correlation between initial performance and long term performance across errors, as well as the presence of further 'adaptation bumps' between epochs 4 and 8. The consistency of this range suggests that encountering a variant maze which causes innovative learning tends to happen, on average, three to four epochs after the initial learning. 

We have, however, observed this range varying for individual cases with some instances experiencing multiple small bumps, and others presenting with one substantial spike to nearly the initial performance level, followed by an on model return to reciprocal behavior. In trials investigating long run trends (extending to 100 epochs), we observed that the average case over each error level achieved asymptotic performance by 9 epochs, with no statistical outlier cases of sudden change in performance level ever occurring after 17 epochs across 1000 instances of training.

\begin{figure}[b]
\centering
\includegraphics[scale = 0.5]{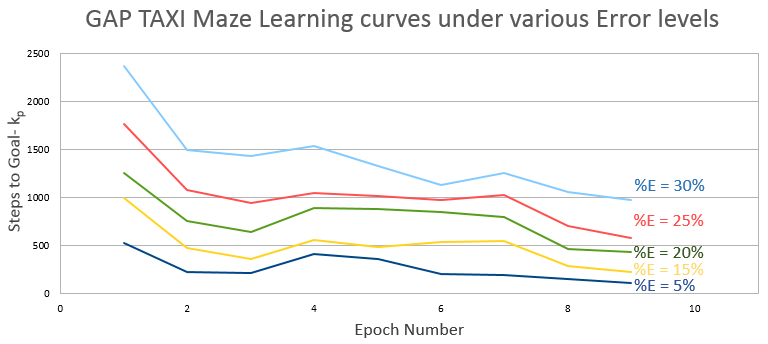}
\caption{Performance of the GAP algorithm across multiple levels of induced error on the Maze/TAXI problem space}
\label{fig:MAZE_TAXI_error_curves}
\end{figure}

\begin{figure}[t]
\centering
\includegraphics[scale = 0.5]{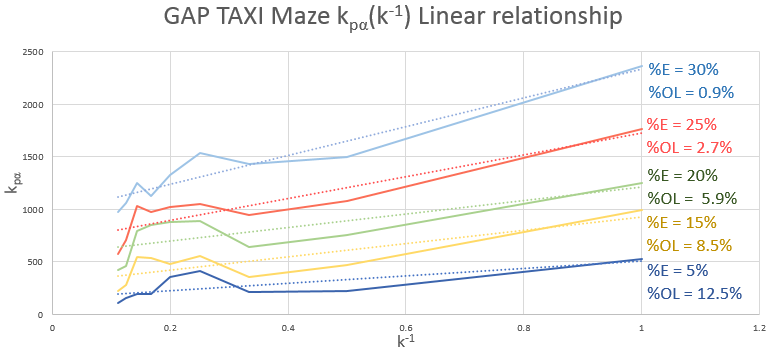}
\caption{Plot of $k_p$ as a function of $\frac{1}{k}$, and the measures of  non-linear divergence for the Maze/TAXI trials at each induced error level.}
\label{fig:MAZE_TAXI_recip}
\end{figure}

Additionally, we want to establish continued adherence to our expected learning curve in preparation for more sophisticated analysis of the data based on these assumptions. To demonstrate this, we have again plotted the relationship between $k_{p\alpha}$ and $\frac{1}{k}$, in Figure \ref{fig:MAZE_TAXI_recip}. We see low levels of off linear error for each curve, with the scale of these errors being roughly inversely proportional to the induced error. This is a manifestation of the same phenomena described in the prior section whereby early convergence skews the fit equation, in this particular case the early skew being due to error rather than abstraction.

We have here explored the performance over error regimes, with some interesting hints as to the capacity for learning transference, but our primary goal is further investigation of the impact of abstractions on the GAP algorithm's learning. The primary purpose of the error panels is that we will be using the method described in Section 5.1.4, based on error thresholding, to identify $L_{max}$ for a group of abstractions used \textit{in combination}.

For this additional battery of experiments, we apply three different kinds of abstractions to the state. These naturally represent reductions in the total expressiveness for the agent's learning system, but when adequate to enable valid solutions, also offer the chance for more rapid convergence of learning. We refer to the three as AI, AII, and wA:

\underline{AI} constructs a vector representing the 8 neighborhood cells to the agent's current cell;

\underline{AII} is similar to AI, but includes only the 4 cells above, below, and to the sides of the current cell; and 

\underline{wA}, or 'with Action', adds the additional information of the most recent action the agent has taken to the full state vector, inspired by the colloquial 'right hand rule' for naive maze navigation. 

\begin{figure}[b]
\centering
\includegraphics[scale = 0.55]{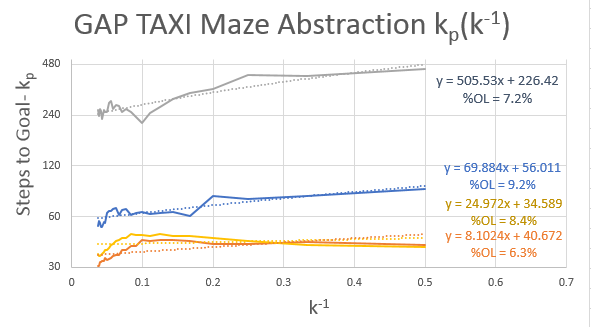}
\caption{Linearized plot of the learning curves for the Maze/TAXI learning under abstraction with corresponding measures of off linear performance for each of the four combinations of abstractions implemented}
\label{fig:maze_taxi_abst_recip}
\end{figure}

We produce four different state generation methods with these: 'AI wA', using AI and wA together, 'AII wA', and AI and AII both without wA (nominally 'AI w/oA' and 'AII w/oA'). By joining the different models in this way, we can compare the relative impact of each different abstraction, in accordance with the form of Equation \ref{eqn:kp_lim_ab}.

\autoref{fig:maze_taxi_abst_recip} illustrates the linearized performance curves across the four joint abstractions, with their off linear errors, of which all are less than 10\%. We can also observe directly that the learning curves of 'AII wA' and 'AII wA' are very similar, with 'AII w/oA' being less effective, but in the same general range, and finally that 'AI w/oA' performing substantially worse, with convergent behavior an order of magnitude worse than the other three. Taken together, this highlights the critical sensitivity of performance to $||T_{\alpha k}||_1$ demonstrated in Equation \ref{eqn:Q_emp} and discussed in Section 5.2, with the inclusion or  non-inclusion of the 'wA' abstraction alone being responsible for a 13-fold reduction in asymptotic performance.

\begin{table}[t]{
\centering
\begin{tabular}{|c|r|c|c|c|c|}
\hline
$P_{Thresh}$ & &  AI wA  & AII wA & AI w/oA & AII w/oA \\
\hline
1\%     & Meas: & 30.19 & 23.40 & 396.17 & 38.00 \\
        & Pred: & 28.27 & 22.42 & 452.29 & 37.19\\
\hline
5\%     & Meas: & 54.16 & 24.58 & 272.13 & 30.88 \\
        & Pred: & 54.71 & 24.86 & 231.61 & 29.66\\
\hline
10\%    & Meas: & 52.49 & 31.72 & 285.07 & 37.76 \\
        & Pred: & 53.39 & 31.69 & 234.67 & 39.68\\
\hline
15\%    & Meas: & 67.30 & 35.43 & 418.00 & 35.83 \\
        & Pred: & 71.19 & 37.33 & 492.38 & 40.62\\
\hline
20\%    & Meas: & 42.12 & 31.66 & 729.29 & 36.64 \\
        & Pred: & 40.83 & 34.89 & 788.94 & 39.97\\
\hline
25\%    & Meas: & 58.45 & 29.70 & 464.91 & 51.31 \\
        & Pred: & 54.78 & 30.39 & 399.60 & 58.65\\
\hline
Avg. \%E: & & 4.03\% & 3.88\% & 14.45\% & 7.98\% \\
\hline
\end{tabular}
\caption{Predicted $k_{p\alpha}$ versus measured $k_p$ across error and abstraction for Maze/TAXI}
}\end{table}

We also use the profile of each joint abstraction across error levels to measure the same indicators of correlation to the model as before, the first set being the measured and predicted $k_p$ laid out on Table 4. On this table, we have each abstraction at each of the error levels sampled, and the average percent error across abstractions. For 'AI wA', 'AII wA', and 'AII w/oA' we have relatively low errors at 4-8\%, with 'AI w/oA' being an outlier at 14.5\% error. In this particular case, we discover that this is the reverse case of the 'early convergence' phenomena, where by extending the runs past the established 10 epoch threshold, we find that the 'AI w/oA' trials were not fully converged until approximately 15 epochs. Though they are then within about 5\% of the value at 10 epochs, this is enough to skew the fit.

\begin{table}[b]{
\centering
\begin{tabular}{|c|c|c|c|}
\hline
  & $L_{max}$ Pred & $L_{max}$ Meas & \%E \\
\hline
AI wA     & 61.7  & 69.1 & 10.6\% \\
\hline
AII wA    & 32.2  & 36.8 & 12.5\% \\
\hline
AI w/oA   & 527.1 & 573.1 & 8.0\% \\
\hline
AII w/oA  & 40.4  & 43.8 & 8.3\% \\
\hline
\end{tabular}
\caption{Comparison of measured and predicted $L_{max}$ across abstractions for the complex Maze/TAXI domain with joint abstractions}
}\end{table}

Table 5 additionally presents the calculated values for $L_{max}$. We find that the pairs of values are within the scale of correspondence observed as typical for the GAP algorithm thus far, and on the appropriate scale for the performance values observed in Table 4.

In addition to these observations, we can also make some substantial insights through the use of induced error in addition to the abstractions, allowing us to calculate estimates of several system parameters by measurement of the proportional effects achieved by the various combinations of abstractions. Because a joint $P_{\alpha}$ can be constructed by multiplying together the constituent abstractions comprising it, and these can be observed in their effect via Equation \ref{eqn:kp_lim_ab}, we can use the discovery of $k_p$ and $L_{max}$, along with the the fit functions for $k_{p\alpha}$ as a function of $log(1-P_{thresh})$, to calculate $||\alpha^+_{Ts}||_1\cdot{}||T_{\alpha k}||_1\cdot{}||\alpha_{Ts}||_1$.

\begin{figure}[t]
\centering
\includegraphics[scale = 0.55]{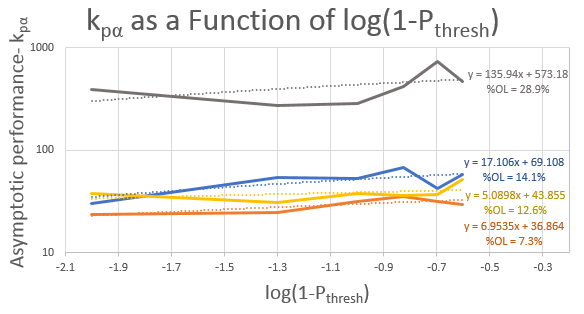}
\caption{Convergent behavior of the four abstractions as a function of $log(1-P_{thresh})$, as in Eq. \ref{eqn:kp_lim_ab}, along with the percent off linear deviations for each curve, corroborating the use of Eq. \ref{eqn:kp_lim_ab} as a proxy for calculating the remaining component $|\alpha^+T\alpha|$}
\label{fig:mt_abst_pthresh}
\end{figure}

\begin{table}[t]{
\centering
\begin{tabular}{|c|c|c|c|c|}
\hline
& \multicolumn{2}{c|}{AI} & \multicolumn{2}{c|}{AII}\\
\hline
$P_{Thresh}$ &  $k_{p\alpha}$  & $|\alpha^+T\alpha|$ & $k_{p\alpha}$ & $|\alpha^+T\alpha|$ \\
\hline
1\% & 30.19 & 1.05 & 23.40 & 1.15 \\ 
5\% & 54.16 & 1.09 & 24.58 & 1.11 \\
10\%& 52.49 & 1.06 & 31.72 & 1.21 \\
15\%& 67.30 & 1.57 & 35.43 & 1.77 \\
20\%& 42.12 & 1.02 & 31.66 & 1.14 \\
25\%& 58.45 & 1.05 & 29.70 & 1.08 \\
\hline
\multicolumn{1}{|c|}{wA}  & \multicolumn{2}{c|}{1.144 ($\pm$12.5\%)} & \multicolumn{2}{c|}{1.249 ($\pm$14.1\%)}\\
\hline
1\% & 396.17 & 1.01 & 38.00 & 0.99 \\ 
5\% & 272.13 & 1.00 & 30.88 & 0.99 \\
10\%& 285.07 & 1.00 & 37.76 & 1.00 \\
15\%& 418.00 & 1.01 & 35.83 & 0.99 \\
20\%& 729.29 & 0.99 & 36.64 & 1.00 \\
25\%& 464.91 & 1.01 & 51.31 & 0.00 \\
\hline
\multicolumn{1}{|c|}{w/oA} & \multicolumn{2}{c|}{ 1.004 ($\pm$0.3\%)} & \multicolumn{2}{c|}{0.996 ($\pm$0.2\%)}\\
\hline
\end{tabular}
\caption{Measured $k_{p\alpha}$ and corresponding $|\alpha^+T\alpha|$ estimates}
}\end{table}

On Figure \ref{fig:mt_abst_pthresh}, we have plotted $k_{p\alpha}$ as a function of $log(1-P_{thresh})$, in order to establish that the correct behavior of Equation \ref{eqn:kp_lim_ab} exists for the succeeding analysis. It is possible to see here that while the error rates for these plots are higher than we have been seeing for the actual learning curves, especially for 'AI w/oA', there is still a qualitatively visible linear pattern, from which reasonable approximations may be made. 

Given this form, we can then calculate the remaining values in Equation \ref{eqn:kp_lim_ab} from the curves in Figure \ref{fig:mt_abst_pthresh}, which yields Table 6, containing the estimated values of the L1 norm of the abstracted transition array (denoted thereupon as $|\alpha^+T\alpha|$ for compactness). Because we can use the variant $k_{p\alpha}$ and $P_{thresh}$ at each induced error level alongside the generally derived $L_{max}$, we are able to produce an estimate for $|\alpha^+T\alpha|$ at each error level. These show a relatively small variance across error level, as expected given that the abstraction matrices are constant, though there is some statistical variance, due in part to the stochastic nature of the experiment trials, and likely also to approximation errors evolving from the fit functions of Figure \ref{fig:mt_abst_pthresh}.

With the values calculated across error levels for $|\alpha^+T\alpha|$ being in close correspondence, we can conclude that we have successfully estimated this parameter. However, it is by nature the estimate for the total abstraction across the mapping from the 'measured' state to the 'true' state, and may contain other factors not due to the specifically controlled mappings, such as other components of the state vector (like the effect of the 'carries passenger' component) or implicit features of the agent's workspace (of the same type as the implicit abstraction in assuming a well conditioned maze, discussed earlier). 

Because we elected to combine the abstractions as groupings of subsets of two, though, we can make an estimate of the impact in transitioning from one of the paired subsets to the other, relying on the submultiplicity of the L1 norm. By taking the ratios of the pairs' measures, we can approximate the impact of each transition from 'AI' to 'AII', both in the 'wA' and the 'w/oA cases, and compare these, and vice versa for 'wA' and 'w/oA' across 'AI' and 'AII' both. 

\begin{table}[b]{
\centering
\begin{tabular}{|c|r|c|c||c|c|c|}
\hline
Q($\alpha$) & AI & AII & $I\rightarrow II$ &  & AIwA$\rightarrow$AIIw/oA\\
\hline
wA      & 1.144 & 1.249 & 1.091 & Meas: & 0.871 \\
\hline
w/oA    & 1.004 & 0.996 & 0.992 & Pred 1: & 0.958 (+10\%)\\
\hline
wA $\rightarrow$ w/oA   & 0.877 & 0.798 &  & Pred 2: & 0.791 (-9\%) \\
\hline
\end{tabular}
\caption{Calculated $|\alpha^+\alpha|$ ratios across abstractions and predicted transform measure, derived from the entries in Table 6 and Eq. \ref{eqn:kp_lim_ab}}
}\end{table}

On Table 7, we have these values, and of specific note is the level of correspondence between the independent transition ratios. When presuming 'AI', changing from 'wA' to 'w/oA' effects a 0.877 factor shift in the L1 norm of the abstracted transition array, and in the 'AII' case, a 0.798 scale change. The transition from 'AI' to 'AII' measures as a 1.091 factor for 'wA', and 0.992 for 'w/oA'. Both sets present extremely similar scale changes, suggesting strongly that they are very close to the actual impacts predicted by Equation \ref{eqn:kp_lim_ab}. It is difficult, however, to judge the scale of these deviations, as a ground truth measure is not directly available. Instead, we can get a sense of the net comparison by independently combining the two transitions in the two combinations which transfer 'AI wA' to 'AII w/oA' and compare these to the actual proportional difference between 'AI wA' and 'AII w/oA'. Doing so, we find the first estimate to be 0.958, and the second 0.791; respectively 10\% and 9\% off of the actual ratio of 0.871, which puts the relative accuracy into proportion.

\subsection{Tower of Hanoi Domain}

The Tower of Hanoi puzzle is a perennial favorite problem class for mathematical analysis. Conceptually simple, the puzzle consists in the most basic form of a number of disks and three or more pegs on which these disks may be stacked, with the disks labeled by an ordinal index which must be preserved when disks are transferred between pegs. The canonical implementation has 3 pegs, and typically from 3 to 7 disks, but both categories can be expanded to change aspects of the problem case. Expanded problems are usually represented as $ToH_{p,d}$, where $p$ is the number of pegs, and $d$ is the number of disks.

\begin{figure}[b]
\centering
\includegraphics[scale = 0.6]{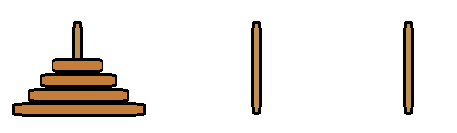}
\caption{Illustration of a traditional Tower of Hanoi (ToH) problem: the objective is to move all disks from the first peg to the third, by only moving disks between pegs, and under the constraint that a disk may only be moved on top of a larger disk or to an empty peg. This graphic shows the 3-peg, 4-disk, variant of the problem, $ToH_{3,4}$}
\label{fig:ToH_prob}
\end{figure}

This problem provides a wide range of benefits for experimentation and demonstration with regards to exploring properties of a problem solving agent. It is a well defined problem which is straightforward to simulate and solve, with the mathematics of the 3-peg case being particularly well studied. This allows for direct performance bounding, with greater numbers of pegs presenting wider state spaces with lower net complexity, and increased numbers of disks representing increases in the depth of complexity.

For our purposes, there is also the added benefit of the state space and action space complexity. The scope of the state space is around $O(p^d)$, depending on whether pegs are considered interchangeable, and the action space at $O(p^2)$. Naturally, in this particular space, only a small subset of the space forms a reachable neighbourhood from any other state, so the graph complexity is lower than the upper bound, but this still represents a substantial space to explore. Further, by contrast to the prior cases which had fairly small action sets regardless of state space size, the Tower of Hanoi problem allows for more expansive action spaces by increased number of pegs.

\begin{figure}[t]
\centering
\includegraphics[scale = 0.45]{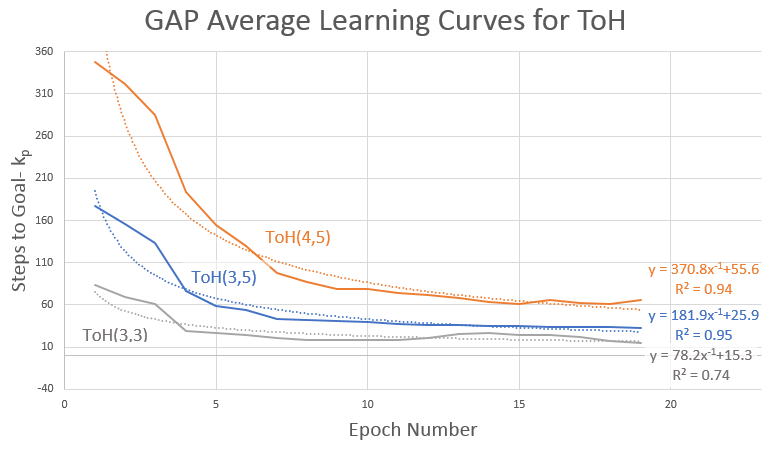}
\caption{Average learning curves for the GAP algorithm over the three investigated ToH domains, $ToH_{3,3}$, $ToH_{3,5}$, and $ToH_{4,5}$ at varying error levels, along with reciprocal fit curves}
\label{fig:ToH_avg_curves}
\end{figure}

Here, we investigate three instances of the problem, in order to explore the effectiveness of learning across multiple complexity classes on a constant problem. Figure \ref{fig:ToH_avg_curves} plots the average learning curves for $ToH_{3,3}$, $ToH_{3,5}$, and $ToH_{4,5}$ over error rates ranging from 0\% to 25\%, and the reciprocal best fit curves for each. As with the prior two domains, we see close fit to the reciprocal form. One interesting observation we can make across varying complexity is that in each case both the scale factor $A$ and the asymptotes $k_p$ increase by approximately a factor of 2, indicating that both $k_p$ and $L_{max}$ double over each increase in complexity, meaning that the learning behavior is essentially identical, but delayed and elevated for larger state/action spaces.

As with the prior cases, we also investigate the performance of the GAP algorithm under changing error levels. Table 8 presents the results from these tests in tabular format. Here, we can see steady deviations from the near optimal performance of the $ToH_{3,3}$ and $ToH_{3,5}$ cases as error level increases. Perhaps more interesting, we can see that the $ToH_{4,5}$ levels deviate substantially with error, suggesting that the expanded state space is more vulnerable to impacts of error (a property we will interrogate further shortly). In addition to the convergent performance values, we also present the errors associated with the reciprocal fit curves, \%OL, showing generally strong fits, excepting the outlier of the $ToH_{3,3}$ case at 5\% error. However, the low error between $k_p$ and $k_{p\alpha}$ suggests that this is likely due to rapid convergence as with several of the previous low error cases, and indeed the $ToH_{3,3}$ case converges at approximately 5 epochs rather than the 20 sampled.

\begin{table}[t]{
\centering
\begin{tabular}{|c|c|c|c|c|c|}
\hline
  & $P_{thresh}$ & $k_p$ & $k_{p\alpha}$ & \%E & \%OL \\
\hline
            & 5\%   & 16.5 & 15.6 & 5.5\% & 19.3\% \\
ToH(3,3)    & 15\%  & 47.3 & 47.8 & 0.9\% & 7.8\% \\
            & 20\%  & 61.9 & 63.4 & 2.4\% & 1.8\% \\
\hline
            & 5\%   & 32.7 & 30.1 & 11.4\% & 5.4\% \\
ToH(3,5)    & 15\%  & 34.8 & 37.1 & 6.4\% & 5.8\% \\
            & 20\%  & 44.9 & 48.7 & 8.2\% & 16.1\% \\
\hline
            & 5\%  & 206.4 & 201.5 & 2.4\% & 12.3\% \\
ToH(4,5)    & 15\% & 696.7 & 707.5 & 1.6\% & 2.3\% \\
            & 20\% & 2052.2 & 2006.9 & 3.0\% & 1.2\% \\
\hline
\end{tabular}
\caption{Chart of the correlation measures for the GAP Algorithm learning the Tower of Hanoi problem, across error level and problem complexity class}
}\end{table}

Because the Tower of Hanoi is such a well studied problem, optimal algorithmic solutions are available from \cite{van1990complexity}, and we can therefor compare the asymptotic performance of the agent to the theoretical optima. For the 3 peg cases, the optimal number of moves is $2^d-1$, giving 7 moves for 3 disks and 31 moves for 5 disks. The convergent behavior for the agents on these cases with random error are 15 and 33 steps respectively across the full error range, with the 0\% error cases naturally achieving the optimal performance level after one epoch.

We can also use the substantial state space, evolving out of a restricted set of elements (the disks, pegs, and rules for movements) for further analysis. This makes the Tower of Hanoi state space readily amenable to developing state abstractions, of which we develop and implement four, referred to as 'AI', 'AII', 'AIII', and 'AIV'. These abstractions are:

\underline{AI}- Direct conversion of lists of disks on each peg to a numerical state: the sum of products of disk indices on each peg;

\underline{AII}- Encoding of disk placement as a list of the sums of disk indices on each peg;

\underline{AIII}- Listing pairs for, each peg, of the number of disks currently stacked and the index of the topmost disk;

\underline{AIV}- Listing the number of disks on each peg;

For example, for the 3-peg, 5-disk problem, if the current state were represented in full as $\{[1,3],[2],[4,5]\}$, AI would produce '25'; AII would give '[4,2,9]'; AIII yields '[(2,1),(1,2),(4,2)]'; and finally AIV has '[2,1,2]'. Each of these represents an increased level of state reduction, as measured by the number of total aliased states, with AI being nearly equivalent to the full state space, and AIV being massively reduced.

\begin{table}[t]{
\centering
\scalebox{0.93}{
\begin{tabular}{|l|l|l|l|l||l|l|l|}
\hline
  & Abst. & $k_p$ & $k_{p \alpha}$ & \%E & $L_{max}$ & $L'_{max}$ &  \%E \\
\hline
            & AI    & 16.5 & 15.6 & 5.6\% & 15.6 & 17.5 & 11.4\% \\
$ToH_{3,3}$ & AII   & 27.8 & 31.5 & 13.5\% & 8.1 & 8.8 & 7.3\% \\
            & AIII  & 21.6 & 17.3 & 20.2\% & 17.22 & 14.8 & 16.3\% \\
            & AIV   & 17.0 & 15.3 & 9.7\% & 31.5 & 35.1 & 10.3\% \\
\hline
            & AI    & 35.3 & 34.2 & 3.1\% & 62.1 & 59.4 & 4.7\% \\
$ToH_{3,5}$ & AII   & 31.4 & 35.1 & 11.8\% & 64.9 & 69.0 & 5.9\% \\
            & AIII  & 31.0 & 31.0 & 0\% & N/A & N/A & N/A \\
            & AIV   & 31.0 & 31.0 & 0\% & N/A & N/A & N/A \\
\hline
            & AI    & 254.5 & 256.9 & 0.9\% & 112.06 & 104.5 & 6.8\% \\
$ToH_{4,5}$ & AII   & 278.1 & 241.3 & 13.2\% & 101.9 & 112.6 & 10.5\% \\
            & AIII  & 1267.2 & 1354.6 & 6.9\% & 391.9 & 371.5 & 5.2\% \\
            & AIV   & 2017.7 & 1843.2 & 8.6\% & 719.9 & 749.8 & 4.1\% \\
\hline
\end{tabular}}
\caption{$k_p$ and $L_{max}$ comparisons for the GAP algorithm learning the ToH problem with various abstractions and across complexity classes, note that for ToH(3,5), AIII and AIV consistently converged to the optimum after one epoch, prevent fit curves being derived to calculate $L_{max}$}}
\end{table}

\begin{table}[b]{
\centering
\scalebox{0.85}{
\begin{tabular}{|c|r|c|c|c|c|c|c|c|c|}
\hline
     & AI &  & AII &   & AII &   & AIV & \\
\hline
                & $A k^{-1}$ & \% OL & $A k^{-1}$ & \% OL &$A k^{-1}$ & \% OL &$A k^{-1}$ & \% OL \\
\hline
$ToH_{3,3}$     & 130.9 & 4.1\% & 84.9 & 6.1\% & 116.9 & 8.0\% & 131.6 & 6.7\% \\
\hline
$ToH_{3,5}$     & 144.1 & 1.6\% & 195.6 & 3.1\% & N/A & -\% & N/A & -\% \\
\hline
$ToH_{4,5}$     & 721.5 & 8.2\% & 439.7 & 6.5\% & 3886.6 & 7.5\% & 989.4 & 1.6\% \\
\hline
\end{tabular}}
\caption{Curve fit metrics for $k_{p\alpha}=\frac{A}{k}+k_p$ in the ToH trials, across Abstractions I-IV, with \% off linear measures for each best fit line.}
}\end{table}

Of note, in addition to these four state compression models, we also attempted trials with one additional abstraction: a reduced form of AIII including only the size of the topmost disk. However, in this case learning convergence failed wholesale, with the 'asymptotic' performance invariably being within a single standard deviation of the initial performance, indicating essentially random action similar to the observed situation when increasing the induced error rate to extreme levels in the Maze/TAXI domain.

Table 9 presents the essential measurements for the entire battery of experiments, spanning $ToH_{3,3}$, $ToH_{3,5}$, and $ToH_{4,5}$; all four abstractions across error rates from 0 to 20\%. A few features of this aggregate data are immediately notable: first, we observe that the AIII and AIV cases for the $ToH_{3,5}$ case unilaterally converge to the optimal number of steps, precluding analytical estimation of $L_{max}$. This is remarkable as this convergence is not strictly seen in the simpler $ToH_{3,3}$ case. Further, the highest performing abstractions vary among the problem classes; for $ToH_{3,3}$ AI and AIV perform most strongly; AIII and AIV for $ToH_{3,5}$; and finally AI and AII for $ToH_{4,5}$. 

Additionally, on Table 10 are presented the coefficients for the reciprocal fit ($A \frac{1}{k} + k_p$) of each trial set for each of the ToH cases, and the corresponding off linear error associated with each such curve. We can see that the curves match the predicted form to comparable levels as in the previous problems, with errors on the order of 2-6\%.

One particular comparison between these pair of tables to be made is in the $ToH_{4,5}$ cases for AIII and AIV. For AIII and AIV, the convergent $k_p$ are substantially larger than those for the AI and AII cases on this same problem. We may hypothesize that this could be due to insufficient numbers of epochs explored to effect learning, however the corresponding fit to reciprocal form have errors of only 7.5\% and 1.6\%, well on par with other experiments not presenting immature learning. Further, we can note that the average, no abstraction $ToH_{4,5}$ curve has convergent behavior at about 56 steps to goal, as evidence that the $ToH_{4,5}$ problem is not inherently resilient to the GAP algorithm's learning, and thus we conclude that these abstractions are therefore ill conditioned to the problem case. 

We can apply similar rationale to the AI and AII cases, which have terminal behavior substantially worse than the  non-abstracted case, unlike the counterparts in the $ToH_{3,3}$ and $ToH_{3,5}$ cases, both of which are well within range of the average case performance. Because all of the abstraction trials using $ToH_{4,5}$ perform substantially worse than the  non-abstracted case, yet the other problem classes perform similarly, we my hypothesize that the cause of the discrepancy is the unsuitability of the AI-AIV models to represent the $ToH_{4,5}$ problem space, specifically.

\begin{figure}[t]
\centering
\includegraphics[scale = 0.42]{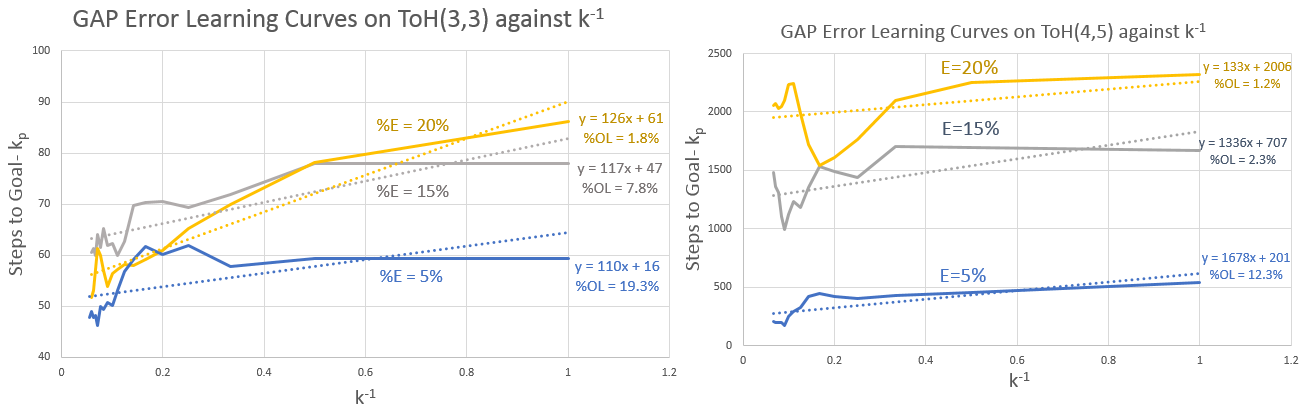}
\caption{Plots of the linearized GAP learning curves for the ToH(3,3) (left) and ToH(4,5) (right) problems across multiple levels of induced error and associated off linear error measures}
\label{fig:ToH_33_45}
\end{figure}

Indeed, if we examine AI in particular, we may note that by increasing the number of pegs, we have created a case in which the number of abstracted states remains nearly constant (as a function of the disk placements) yet the number of real states has increased exponentially. In a general sense, we can consider this ratio of abstracted to real states as a metric component which puts an upper bound on $||T_{\alpha k}||_1$, and by extension $Q(\alpha_T)$. Similar complications thus exist at even more substantial levels for the other abstractions, with more severe impacts due to the greater reduction in the size of the abstracted state space.

One way we can see this effect in action is by using error rate as a proxy for effectiveness on performance. In Figure \ref{fig:ToH_33_45} (left) are plotted the linearized curves for $ToH_{3,3}$ across induced error rates 5, 15, and 20\%. Here, we see that the $k_p$ associated with the abstractions congregate around the 5\% and 15\% error levels, and with a ratio of 3.8x between the 20\% and 5\% $k_p$. In Figure \ref{fig:ToH_33_45} (right), by contrast, are the error impacts on the $ToH_{4,5}$ case. Here, we can first see that there is a substantially higher susceptibility to induced error for this problem case, with the proportional change between 5 and 20\% being a factor of 10x. In this case, we can see that the AI and AII cases, with $k_p$ around 260, are near to the 5\% error level, with the AIV case approaching the 20\% error level. 

By these comparisons, we can see the ways in which the relative size of an abstracted state space impacts overall performance, producing similar results to that of changing error level, as predicted by the existence of the $Q(\alpha_T)$ metric. The implication is that, as we can measure overall performance in terms of $k_p$, and there is a direct relationship between $||T_{\alpha}||_1$ and $k_{p \alpha}$, the impact of the ratio of the sizes of the native and abstracted state spaces has a substantial limiting impact on the overall performance of the GAP algorithm.

\section{Conclusions and Future Work}

In this paper, we have presented a hypergraph based machine learning algorithm, the Goal Agnostic Planner, or GAP, designed to learn hierarchical planning problems. This algorithm uses a 3-dimensional array, modeling a hypergraph data structure and augmented by two 2-dimensional composite data structures comprised of arrays containing ordered linked lists. These data structures are used to retain information pertaining to occasions, or state-to-state transitions precipitated by actions. We claim that this structure, and the associated algorithms, provide our machine learning and planning system four valuable properties: Goal Agnosticism(I); Algorithmic Efficacy(II); Abstraction robustness(III); and Learning Convergence(IV).

The additional array/linked lists are used to model 2-dimensional slices of the hypergraph, and we proved that these slices contained the path through the state space with the greatest joint probability path between any pair of states embedded in the hypergraph. We also developed, to accompany this data structure, a maintenance algorithm which supports the ordering and structure of the augmentation arrays. We showed that during the modification of the primary hypergraph with new recorded observations this optimal path property is preserved, underwriting claim II. We then presented a sequence inference algorithm based on Dijkstra's algorithm which treats the 2D array/linked lists as graph slices of the higher dimensional hypergraph, and proved that this algorithm extract the greatest probability path in the hypergraph between any pair of states, supporting claim I.

We then used this information to construct a model of dynamic agent performance by converting the maximal probability tree associated with the hypergraph slices into a transition matrix for Markov chain analysis. Using this modeling approach, in conjunction with the observation that we can model any goal state as an attractor state, we are able to produce relationships describing the time dependent probability of transition from arbitrary states to the goal state. We also derived the conditions under which a system may be incapable of transitioning to the goal (trap nets), and time bounds on stochastic transition to the goal state.

Based on these relationships, we were able to introduce a model for abstraction as a transform between two transition arrays. Using this transform, we are able to derive the conditions under which a path planned in the abstracted space will also be a valid path in the true state space, demonstrating property III. We also analyzed the impact of the abstraction on system performance, and derived a correlated metric for describing the 'quality' of an abstraction in terms of this performance change. We used a specialized one-to-one transform to model of incremental learning, showing that this transform approaches the identity as learning progresses, proving property IV. Additionally, from the successive analysis we were able to determine that the learning curves will, on average, follow a reciprocal trend.

To investigate the performance of the GAP algorithm on actual problem cases, we performed trials on three problem cases: a fixed environment procedural task based on traditional problem cases explored by the STRIPS algorithm; a variable environment problem constructed by combining maze navigation and the TAXI domain; and the Tower of Hanoi puzzle under multiple configurations. In each case, we examined the performance of the GAP algorithm during learning, demonstrating that the predicted reciprocal form of the convergent learning curve persists throughout all experiments.

To validate other components of our analysis, we made proxy measurements derived from the best fit learning curves to compare to values predicted based on our analysis and calculated the correspondence between the measured and predicted values. We further applied varying levels of artificial error in each experiment case to study the impact on learning performance, showcasing persistent convergent behavior in the face of these disturbances. We also used the increased complexity and larger state spaces of the Maze/TAXI and Tower of Hanoi problems to investigate the effects of various abstractions on learning performance.

Aside from these broad investigations, we also used individual experiments to explore specific properties related to the GAP algorithm. We use the STRIPS problem case to study the power law relationships predicted between convergent performance under varying error rates. For the Maze/TAXI case, we implemented a set of abstractions which could be applied in tandem, using the derived performance relationships to confirm theoretical predictions about abstraction transforms under composition. We then used error induction as a means to validate the relationship between performance under abstraction and learning rate thresholds established prior. In the Tower of Hanoi experiments, we used variant abstractions of increasing levels of state compression to investigate the properties of the abstraction quality metric, comparing it to performance changes induced by artificial error, and further developed the understanding of the quality metric using these relationships.

In summation, experimental cases were used to show applied examples of all four properties, as well as providing additional measurements to confirm the predictions of the analysis. Additionally, relationships between different complexity classes within individual domains were used to further explore some qualitative aspects of the GAP algorithm's behavior.

\subsection{Limitations}

In this section, we discuss some of the outstanding limitations associated with implementation of the GAP algorithm.

\subsubsection{Memory Use of INC Array}

Perhaps the most substantial hurdle is the size of the INC array, which scales as $O(|S|\times|S|\times|A|)$. Suppose, for a hierarchical problem with two divisions, we can represent each subproblem space by a vector. If each vector contains three binary members, then the total size of the state space will be 64, and thus the state/state space will be of size 4096 elements, multiplied by however many actions are present, highlighting the ease with which the hypergraph can become very large.

One way we might look at this is as an extension of problem (2) from \cite{jimenez2012review}, wherein addressing the issue of combinatorial explosion of paths through a simple state space, we have shifted the computational burden to memory space rather than computation space. 

Abstractions naturally provide a means to reduce this space, as well as the potential for traditional methods of compression. Further, because many such spaces will be highly sparse, it is also possible to represent INC using an alternative graph like structure which does not retain empty regions of the array, such as a constructed hypergraph link object with per node addressing encoded by a hash function of the input $(s_i,s_f,a_l)$ coordinates.

Additionally, alternative methods of representing INC which can reduce the size of the array are discussed in Appendices A.3 and A.4.

\subsubsection{Speed of Dijkstra's Algorithm}

In our application case, we have used a naive implementation of Dijkstra's algorithm which operates on the time order of $O(|S|^{2})$. While this is not substantially weak computational performance, it is, in current form, designed mainly to interact in a natural, easy to understand way with the INC and AFI datastructures. However, more advanced, faster path finding algorithms can be applied to improve overall performance. In Appendix A.4, a model for integration with the A* algorithm is presented as a means for adapting the GAP algorithm to use a faster planning approach.

\subsubsection{Familiarization Phase}

For any problem implementation, the algorithm experiences a period associated with primarily random actions for exploration of the state space, using the initialized random actions in the untrained AFI array.

Such a phase may be programmed as a training component for purposes of shortening training time, on basis of  non-execution of the planning algorithm until a sensible portion of the state space is known reachable. However, even if this is not the case, systems will have a period of fully random exploration until some paths to the goal have been identified. While this is a feature of most learning systems, it is not an ideal situation and patterns randomly expressed during this phase may affect later learning as well. 

Approaches to examining the impact of, and amelioration for, this phenomena are discussed in Appendices A.1, A.2, an A.3. Additionally, a method which we explored in pilot experiments for the Maze/TAXI domain and the Tower of Hanoi domain is the use of Tabu search in the exploration phase. In these initial experiments, we implemented a policy by which the selection of exploration actions was guided by balancing previously explored actions-- that is, rather than taking an entirely random actions when the goal state had not yet been discovered, each time a state was visited the selection of action was made among the heretofore least chosen actions, ensuring that each state was uniformly explored during the familiarization phase. 

\subsection{Further Work}

Though the problem cases explored herein present ample evidence pertaining to the proposed properties of the GAP algorithm, there are a selection of specific weaknesses which are not directly addressed by the current experiments that we would like to examine further.

Firstly, though we developed means to detect and quantify the risk associated with trap nets, none of the problem cases in this paper contain any such networks. Though we can by our analytic devices confirm that such do not present a barrier to operation in these problems, we are unable to directly confirm our predictions pertaining to them. Additionally, though the concept of multiple attractor states (including multiple concurrent goal states) is discussed, all systems demonstrated here possess singular goal states, and thus the impact on performance curves due to multiple target states is not addressed. A related set of experiments we wish to undertake includes the evaluation of learning in the presence of multiple goal states (alluded to in 4.2.1) but unaddressed in the included experiments (though two minimally investigated cased in Appendix B.1 and B.2 illustrate learning in spaces with multiple goals).

Additionally, we do thoroughly explore  non-deterministic action of the GAP algorithm vis-a-vis error induction, as well as by the exploration of abstractions which reduce the size of the state space (causing certain states to be mixed in the learning structure). Such  non-determinism is not strictly uniform across the state/action space as it is random across actions, and thus creates asymmetric distributions depending on the possible outcomes of those actions. However, none of the problem cases presented here is \textit{inherently}  non-deterministic. Though the combination of abstractions with error is likely to produce systems which appear to the agent to be so, it still bears direct investigation to establish performance of these systems on top of inherent uncertainty, rather than uncertainty modeled as a perturbation or a transform of the state space.

A further complex issue is learning transference. In the randomized worlds used in the Maze/TAXI experiments using relativistic states, we have provided some evidence of transferred learning as the changing maze structure presents a variant problem of the same type and complexity, but with diverging state spaces. However, we have not constructed an explicit model describing performance under these conditions beyond considering the adapted learning as a special case of the general form for learning. This condition was alluded to in discussion (within Section 5.3.2) pertaining to the 'adaptation bumps', however we address it only in the context of the learning model, not an explicit adaptation model. The value of a model for transference learning beyond this is visible, however, from the observed consistency in the rate of appearance of these outlier bumps.

One final concept which we wish to investigate further is unsupervised operation in the context of the GAP algorithm. Though the actual learning process for the GAP method can be considered inherently unsupervised, the planning phase itself is not. While out of scope for the matter of this paper, it is simple to conceptualize a model in which the goal is not expressly a singular or set of states, but rather a metric function of some kind. Though this does re-introduce the issues associated with the use of bespoke objective functions, the potential for the path planning algorithm to terminate when identifying a destination state bearing some flavor of property expressed functionally is something of a hybrid model. The potential for learning relationships determining improvement in some quality functions opens the possibility of both unsupervised planning and learning as well as process improvement.

\appendix

\section{GAP Algorithm Modifications}

Aside from the direct investigations in the paper proper, the nature of the GAP algorithm presents itself naturally to a set of direct modifications, all of which may present advantages to alternative problem formulations. We have experimented with each of these in our test cases, but none in sufficient detail as to warrant such expansion of the scope of this paper without more thorough analytic investigation.

\subsection{Implicit learning rate}

Though the GAP algorithm does not incorporate an explicit learning rate parameter, we have shown that there is learning convergence which will be bound by a reciprocal function. This function was intimately tied to the number of recorded observations contributing to the $INC$ array. In the prior sections, we examined this process as an incremental change to a transform on the transition array during learning, but it could also be viewed as an averaging function over the number of samples.

For example, consider the impact of one fluke observation at different times. Presume that we have observed a state-to-state transition nine times, such that:
$$\sum_{\forall k } INC[s_i,s_j,a_l] = 9$$

and that all such observations have been precipitated with action $a_1$ thus far, but the tenth observation is precipitated by $a_2$; the associated a posteriori probabilties, then were previously:
$$P(a_1(s_i)\rightarrow s_j) = 1.0; P(a_2(s_i)\rightarrow s_j) = 0.0$$

whereas after, they are:
$$P(a_1(s_i)\rightarrow s_j) = 0.9; P(a_2(s_i)\rightarrow s_j) = 0.1$$

a relative shift of $10\%$. If the total observations, however, were 99 prior to the anomalous result, then the probability shift would be only $1\%$. This tidily illustrates the way in which learning is tied to the reciprocal function, and how this function bears some learning inertia. If an anomalous result appears early in training, it will take longer to for the agent to learn corrections than if it were to occur later. Further, if learning is performed online with planning, this may bias the agent from exploring certain paths, whereas an offline learning phase would ameliorate this.

An alternative method which can be employed towards online training is the use of an artificial learning rate, implemented as a fixed proportion moving average calculation for the probabilities. Using such a technique, the probabilities at each update are calculated as proportions of a fixed number of samples, with each observation's change in increment calculated as not one instance, but a number proportional to the fixed association of the change in probability. This would calculated by setting the count increment as the scaled proportion between the actual total sum of observations to the desired fixed window, and effecting the change in INC proportional to this scale.

\subsection{Alternative choice planning}

In certain situations, we might consider an agent which, rather than uniformly selecting the most probable action choice at any given state, may instead select from all available actions based on a weighted probability expectation of each.

For instance, in state $s_i$, we have $AFI[s_i,:,:]$ describing the set of actions and results available. For any action/result pair, then, we can an assign a local probability: $<P_{s_i,a_l}(s_f)> = P(a_l|AFI)P_{a_l(s_i)\rightarrow s_f}$ Where $P(a_l|AFI)$ represents the probability that $a_l$ is chosen: the output of the decision algorithm. Using this expression, we can write an alternative method of hypergraph compression based around the probability each state transition:

$$P(s_i \rightarrow s_f) = \Sigma_{\forall a_l} P(a_l|AFI)P_{a_l(s_i)\rightarrow s_f}$$

This equation compresses the hypergraph along the action slice by coupling all action results together with the $P(a_l|AFI)$ function. Note that if we define:

\[ P(a_l|AFI) =    \left\{
\begin{array}{ll}
      1 & a_l = \underset{k}{\operatorname{argmax}}P(a_l(s_i)\rightarrow s_f)\\
      0 & otherwise\\
\end{array} 
\right. \]

Then the compression resolves to the maximally probable subgraph (essentially, a binary definition of an objective function, for which the GAP algorithm as presented herein is a special case). As an alternative, however, consider that we let 
\begin{equation}
P(a_l|AFI) = \frac{P(a_l(s_i)\rightarrow s_f)}{\Sigma_{\forall a_l}P(s_f|s_i,a_l)}
\label{eqn:alt_pol}
\end{equation}

In this case, then, the probability of taking action $a_l$ is proportional to the relative likelihood of $a_l$ resulting in $s_f$ relative to other actions. Such a function could still be used to construct the necessary Markov decision process for analysis, albeit substantially more complex. One might then expect such a system to converge, though less aggressively, but perhaps to also exhibit over all improved asymptotic behavior due to more expansive state space exploration. An envelope function using Equation \ref{eqn:alt_pol} as the input parameter may then enable selection of policy to favor exploitation or exploration as an explicit, bounded system property. This adjustment is similar to the Tabu exploration described in 6.1.3. One might even theorize that it is possible, using this form of construction, to reverse engineer a Q-learning table from the learned world model by using the proposed reward functions and the a posteriori probability function to infer reward as a function of action choice.

\subsection{In situ transfer functions}

One alternative to learning a full $AFI$ table for a problem would be to represent the output of the table, namely the conditional occasion probability, with a modeled function. Such functions are in fact fairly commonplace, as every state machine or combinatorial planner is, in essence, a distilled transfer function for a given problem operating on a fully binary state transition space.

In this paradigm, some function $I(s_i,a_l)$ would either produce a statistical distribution over $s_f$, or $I(s_i,s_f)$ a distribution over $a_l$. Such a distribution might be, for instance, a rule which eliminates potential actions, such as that for a  non-movement action only producing states which possess the same physical location as $s_i$. As in practice full states are often represented with vectors of individual substates, this would amount to storing some portion of the state vector in INC, and then expanded into multiple vectors with latter portions generated by $I$ at each step of the planning algorithm.

In application, this function would then be called during the planning stage, during which only allowable transitions would be evaluated at each step in building the most probable path subtree within AFI, and possibly even further specifying the associated probabilities with all or a subset of these paths vis-a-vis generating entries in INC, allowing for the array to be partly learned and partly generated by $I(\cdot{})$. In this way, the size of the INC space may be reduced, supplanting learned relations for occasions in part with known rules, either written in advance or derived from analysis of a partially learned INC array.

If such a system is available, then through each step in which the GAP algorithm evaluates the transition probabilities in the boundary, rather than fetching from an array, a span of outputs from the transfer function may be collected instead. Naturally, in this case the consistent maintenance of the maximum probability subgraph is likely to be inefficient without the use of problem specific features of the state space. We can, however, note that because each node on the maximal probability tree, as it is built, contains the prior state information leading from the source node, and thus we may instead maintain a single array linked list containing the sorted boundary nodes.

Under this conception, one would essentially be substituting portions of the INC array with pre-existing model knowledge by calculating the transition probabilities from the model where applicable. For instance, in the Maze/TAXI problem case, one could define a simple function describing the effect of move actions based on the known local topology of the maze. This would allow the learning component to be more directed towards identification of the patterns in locating the passengers and destinations, in effect such systems would learning as though $k$ were already advanced, as portions of the INC array would already be 'filled in', as it were. If analysis of INC is used to derive $I(\cdot{})$, then the action of this evaluation would be tantamount to autonomous hierarchical decomposition of the state space.

\subsection{Generalized Heuristics for A*}

In Section 3, mention was made of the use of A* for implementing more computationally efficient versions of Algorithm 2. While the definition of heuristics which would be applicable to any possible problem space would be difficult at best, it is possible identify such an approach, based in the algorithm's construction, rather than the problem space. 

\begin{figure}[t]
\centering
\includegraphics[scale = 0.4]{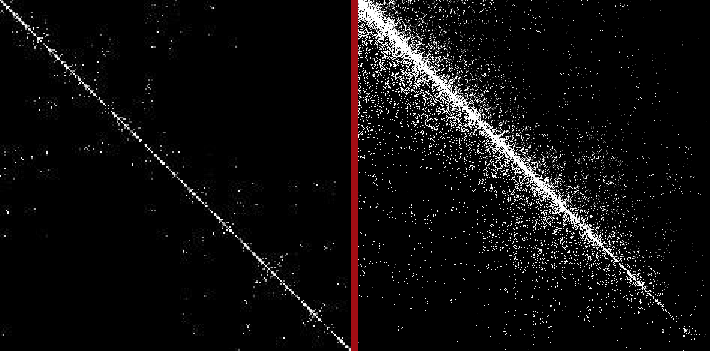}
\caption{Visual representation of example AFI arrays for learning on the $ToH_{3,5}$ problem (left), and the Maze/TAXI problem (right) showing the relationship between structure in the AFI array and state discovery order}
\label{fig:ToH_35_AFI}
\end{figure}

We can illustrate this with a learned structure based example. Recall that we implemented state encoding by use of a hash function, assigning states numerical labels in the order of discovery of the state. Given that, under random exploration, we expect states to be assigned label values roughly in proportion to the number of actions needed to transition between them, we can expect a structural relationship to be present in the AFI graph, whereby transition probabilities are clustered around the primary diagonal. 

We can observe this behavior in practice by examining visual plots of the AFI arrays. Figure \ref{fig:ToH_35_AFI} demonstrates this exact phenomenon for one of the $ToH_{3,5}$ agents and one of the Maze/TAXI agents after learning. Because adjacent states are most likely to be determined within relatively short time periods, there is a tendency towards a statistical distribution around the diagonal, with a measure of state density variance off the diagonal skewing the span of items away from this line, inversely proportional to the state index parameter.

Drawing from this observation, and presuming we have a roughly Gaussian distribution of probabilities around the line $s_i = s_f$ with a variance which decreases approximately linearly from some initial value $v_1$ (measured empirically) as $s_i \rightarrow |S|$, we can use a bit of geometric inference to model the probability estimate $P_e(s_i,s_f)$:

$$P_e(s_i,s_f) \approx \frac{1}{v_1\sqrt{2\pi}}\cdot{}
\frac{2|S|}{2|S|-(s_f+s_i)}\cdot{}
e^{\frac{-1}{v_1}\cdot{}\left(\frac{|S|(s_f-s_i)}{2|S| - (s_f+s_i)}\right)^2}$$

Which approximates the joint probability of transition between two states, extrapolating from distribution of the observed direct transitions. We can then write a heuristic function $h(s_i)$ as $h(s_i) = P_e(s_i,s_g)$, with the starting state being $s_0$, such that A* maximizes:

$$f(s_i) = P_e(s_i,s_g)\cdot{} \prod_{\forall j \in \sigma_{0,i}}  P(s_j\rightarrow s_{j+1})$$

Which function represents the expected total probability from the initial state, $s_0$, to the goal $s_g$, in terms of the actual joint probability from $s_0$ to $s_i$ and the predicted remaining probability of transitioning from $s_i$ to $s_g$ extrapolated from the structure of AFI vis-a-vis the state discovery mechanism.

It is of course important to note that not all problems will allow this sort of adjacent state representation, and thus the problem topology will impact the exact probability distribution used to determine $P(s_i,s_f)$. For instance, in the left of Figure \ref{fig:ToH_35_AFI} we can observe two 'blocks' of available transitions in the upper left and lower right of the array, roughly corresponding to the two phase nature of the workspace where the bulk of the disk are on the first or third peg. The form is much better preserved for the Maze/TAXI case, and in general we would expect the state discovery oriented shape to be more preserved in highly localized problems. However, it is entirely possible to conceptualize a similar process of identifying a statistical distribution over AFI which is not based on the state discovery process which may inform the same probabilistic construction for $h(s_i)$ as illustrated above.

Further, the methods described above as 'In situ transfer functions', may provide a convenient means of developing a model AFI on which this heuristic method may have an alternative distribution generated as well. Indeed, looking at the block like structure in Figure \ref{fig:ToH_35_AFI} suggests that statistical analysis of AFI may allow for autonomously identifiable hierarchical decomposition in INC, as described in A.3, which may allow for the creation of beneficial heuristics or abstractions to represent the state space in more compact form. This approach may provide a design mechanism to analytically reduce the state space size without introducing substantial compression loss, or with probabilistically bounded loss rates.

\section{Additional Experimental Cases}

In this section, we present some minor results from common test cases which are frequently used to illustrate the effectiveness of machine learning and automated planning systems, but which we do not include in the main body of the paper because: (a) they are insufficiently complex to present nuanced analysis investigating sophisticated behavior of the GAP algorithm and (b) the aggressive learning of the GAP algorithm eliminates fine differences between uncertain cases and thus precludes detailed analysis, similar to that seen with the learning of the $ToH(3,5)$ case in Section 5.4.

\subsection{Blocksworld}

\begin{figure}[b]
\centering
\includegraphics[scale = 0.45]{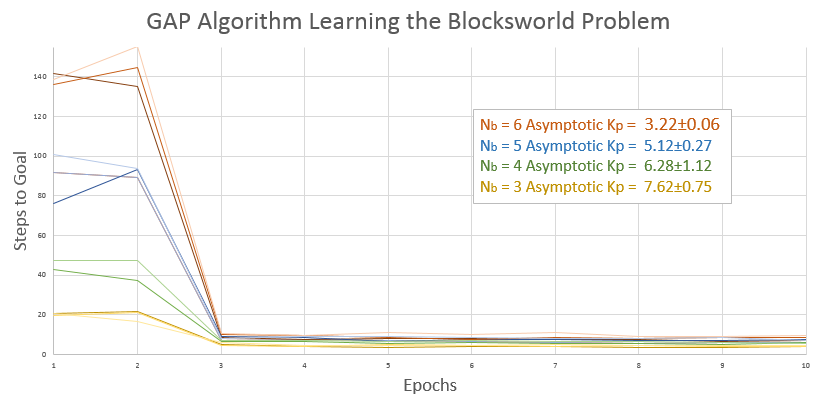}
\caption{Learning curves for the Blocksworld problem, across varying numbers of blocks and error rates. Note that the rapid convergence precludes fitting of a proper reciprocal curve.}
\label{fig:sussman}
\end{figure}

The Blocksworld problem domain is a simple, illustrative example used to study planning algorithms. It consist of, in essence, a 'table' on which blocks can be placed, blocks which can be moved one at a time and stacked, and a goal final state of the blocks stacked on top of one another in order. It is particularly notable for the presence of the Sussman Anomaly, first presented by \cite{sussman1973computational}, a fault present in some kinds of planning algorithms, in which an agent incorrectly resolves interleaved subgoals without generating a valid solution. In essence-- the planning conditions result in the creation of a trap net should continued execution to achieve the goal be performed subsequent to the incorrect problem resolution steps.

We test the GAP algorithm on the Blocksworld problem, with numbers of blocks, $N_b$, ranging from 3 to 6, and at induced error rates of 0, 10, and 20 \%. Each combination of $N_b$ and error rate is sampled across 100 trials of randomly selected initial orientation, running to 10 epochs. For this problem, the proper ordering of blocks, from highest index to lowest, in any position, is considered a goal, and planning is performed by building the maximal probability subtree until any such state is reached.

We find that the GAP system indeed successfully learns the problem, and does not become trapped. Further, as we can observe in Figure \ref{fig:sussman}, the learning process is both expedient and comprehensive, even with induced error rates of 20\%, with convergence to $k_p$ uniformly occurring within 3 epochs. By examination, we find that among our random samples, for $N_b = 3$, 17 of the tests possessed the conditions for the Sussman anomaly to be present, 10 of the $N_b = 4$ trials, 12 of the $N_b = 5$, and 24 for $N_b = 6$, indicating that the GAP algorithm successfully avoids falling prey to the Sussman anomaly in both learning and planning phases.

\subsection{Binary Addition}

Binary addition is the process of, essentially, taking the sequential digits of a pair of binary numbers and attempting to generate the corresponding next pair of digits in their sum. It presents a useful demonstration case for learning algorithms, as it is simple to implement, check, and design reward functions for. In the process of developing the GAP software, binary addition was one of the trial cases we used to check and debug the system. The problem is exceptionally simple, and thus presents little in the way of investigative power, but none the less acts as a fundamental validation of effectiveness.

\begin{figure}[b]
\centering
\includegraphics[scale = 0.6]{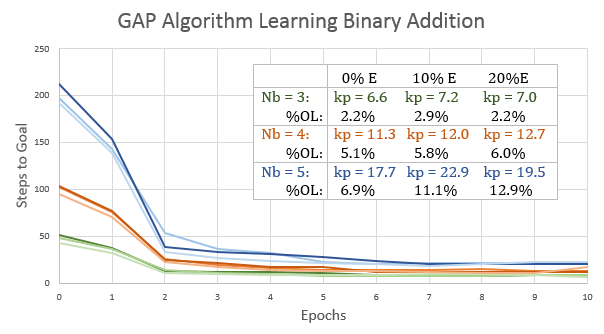}
\caption{Learning curves for the binary addition problem, across 3, 4, and 5 digit numbers and error rates of 0, 10, and 20\%, along with convergent limits for each battery and corresponding off linear percents against reciprocal fit.}
\label{fig:bin_add}
\end{figure}

For this problem, the world consists of two randomly chosen binary numbers of $N_b$ many digits to be added together, a 'carry bit' state, the resultant number, and the actual sum of the former pair, as well as an index to digits of the resultant. The agent's possible actions are toggling the current resultant bit, toggling the carry bit, and incrementing or decrementing the index. States are constructed from the digits at the current index, the carry bit, and the current number of correct digits in the resultant, in a relativistic construction similar to that used for the complex Maze/TAXI problem. The goal states, then, are any state in which all digits are correctly selected. As with the Blocksworld case, we perform experiments in batteries of 100 trials, in this case of 10 epochs each, with errors of 0, 10, and 20\% induced error and $N_b$ of 3, 4, and 5 digit numbers.

Results of these experiments are presented in Figure \ref{fig:bin_add}. On this graph, we can see the learning performance following the expected learning curves, with percentage off linear values and convergent performance limits listed on the inset chart. Notably, across each level of induced error, we can see that the asymptotic performance approximately doubles as $N_b$ increases by one, corresponding to the doubling of the workspace size each time a digit is added.

\bibliography{main}

\begin{thebibliography}{41}
\providecommand{\natexlab}[1]{#1}
\providecommand{\url}[1]{\texttt{#1}}
\expandafter\ifx\csname urlstyle\endcsname\relax
  \providecommand{\doi}[1]{doi: #1}\else
  \providecommand{\doi}{doi: \begingroup \urlstyle{rm}\Url}\fi

\bibitem[Baird and Moore(1999)]{baird1999gradient}
Leemon Baird and Andrew~W Moore.
\newblock Gradient descent for general reinforcement learning.
\newblock \emph{Advances in neural information processing systems}, pages
  968--974, 1999.

\bibitem[Blum and Furst(1997)]{blum1997fast}
Avrim~L Blum and Merrick~L Furst.
\newblock Fast planning through planning graph analysis.
\newblock \emph{Artificial intelligence}, 90\penalty0 (1-2):\penalty0 281--300,
  1997.

\bibitem[Blum and Langford(1999)]{blum1999probabilistic}
Avrim~L Blum and John~C Langford.
\newblock Probabilistic planning in the graphplan framework.
\newblock In \emph{European Conference on Planning}, pages 319--332. Springer,
  1999.

\bibitem[Bylander(1996)]{bylander1996probabilistic}
Tom Bylander.
\newblock A probabilistic analysis of prepositional strips planning.
\newblock \emph{Artificial Intelligence}, 81\penalty0 (1-2):\penalty0 241--271,
  1996.

\bibitem[Dicken and Levine(2010)]{dicken2010applying}
Luke Dicken and John Levine.
\newblock Applying clustering techniques to reduce complexity in automated
  planning domains.
\newblock In \emph{International Conference on Intelligent Data Engineering and
  Automated Learning}, pages 186--193. Springer, 2010.

\bibitem[Dietterich(1999)]{dietterich1999state}
Thomas~G Dietterich.
\newblock State abstraction in maxq hierarchical reinforcement learning.
\newblock \emph{arXiv preprint cs/9905015}, 1999.

\bibitem[Dijkstra et~al.(1959)]{dijkstra1959note}
Edsger~W Dijkstra et~al.
\newblock A note on two problems in connexion with graphs.
\newblock \emph{Numerische mathematik}, 1\penalty0 (1):\penalty0 269--271,
  1959.

\bibitem[Dimitrov and Morton(2009)]{dimitrov2009combinatorial}
Nedialko~B Dimitrov and David~P Morton.
\newblock Combinatorial design of a stochastic markov decision process.
\newblock In \emph{Operations Research and Cyber-Infrastructure}, pages
  167--193. Springer, 2009.

\bibitem[Ding et~al.(2014)Ding, Smith, Belta, and Rus]{ding2014optimal}
Xuchu Ding, Stephen~L Smith, Calin Belta, and Daniela Rus.
\newblock Optimal control of markov decision processes with linear temporal
  logic constraints.
\newblock \emph{IEEE Transactions on Automatic Control}, 59\penalty0
  (5):\penalty0 1244--1257, 2014.

\bibitem[Fikes and Nilsson(1971)]{fikes1971strips}
Richard~E Fikes and Nils~J Nilsson.
\newblock Strips: A new approach to the application of theorem proving to
  problem solving.
\newblock \emph{Artificial intelligence}, 2\penalty0 (3-4):\penalty0 189--208,
  1971.

\bibitem[Floriano et~al.(2019)Floriano, Borges, and
  Ferreira]{floriano2019planning}
Bruno Floriano, Geovany~A Borges, and Henrique Ferreira.
\newblock Planning for decentralized formation flight of uav fleets in
  uncertain environments with dec-pomdp.
\newblock In \emph{2019 International Conference on Unmanned Aircraft Systems
  (ICUAS)}, pages 563--568. IEEE, 2019.

\bibitem[Fox and Long(2003)]{fox2003pddl2}
Maria Fox and Derek Long.
\newblock Pddl2. 1: An extension to pddl for expressing temporal planning
  domains.
\newblock \emph{Journal of artificial intelligence research}, 20:\penalty0
  61--124, 2003.

\bibitem[Fox et~al.(2001)Fox, Barbuceanu, and Teigen]{fox2001agent}
Mark~S Fox, Mihai Barbuceanu, and Rune Teigen.
\newblock Agent-oriented supply-chain management.
\newblock In \emph{Information-based manufacturing}, pages 81--104. Springer,
  2001.

\bibitem[Geffner(2000)]{geffner2000functional}
H{\'e}ctor Geffner.
\newblock Functional strips: a more flexible language for planning and problem
  solving.
\newblock In \emph{Logic-based artificial intelligence}, pages 187--209.
  Springer, 2000.

\bibitem[Georgievski and Aiello(2014)]{georgievski2014overview}
Ilche Georgievski and Marco Aiello.
\newblock An overview of hierarchical task network planning.
\newblock \emph{arXiv preprint arXiv:1403.7426}, 2014.

\bibitem[Grzes(2017)]{grzes2017reward}
Marek Grzes.
\newblock Reward shaping in episodic reinforcement learning.
\newblock 2017.

\bibitem[Jim{\'e}nez et~al.(2012)Jim{\'e}nez, De~La~Rosa, Fern{\'a}ndez,
  Fern{\'a}ndez, and Borrajo]{jimenez2012review}
Sergio Jim{\'e}nez, Tom{\'a}s De~La~Rosa, Susana Fern{\'a}ndez, Fernando
  Fern{\'a}ndez, and Daniel Borrajo.
\newblock A review of machine learning for automated planning.
\newblock \emph{The Knowledge Engineering Review}, 27\penalty0 (4):\penalty0
  433--467, 2012.

\bibitem[Kaelbling et~al.(1996)Kaelbling, Littman, and
  Moore]{kaelbling1996reinforcement}
Leslie~Pack Kaelbling, Michael~L Littman, and Andrew~W Moore.
\newblock Reinforcement learning: A survey.
\newblock \emph{Journal of artificial intelligence research}, 4:\penalty0
  237--285, 1996.

\bibitem[Karami et~al.(2009)Karami, Jeanpierre, and
  Mouaddib]{karami2009partially}
Abir-Beatrice Karami, Laurent Jeanpierre, and Abdel-Illah Mouaddib.
\newblock Partially observable markov decision process for managing robot
  collaboration with human.
\newblock In \emph{2009 21st IEEE International Conference on Tools with
  Artificial Intelligence}, pages 518--521. IEEE, 2009.

\bibitem[Knoblock(1990)]{knoblock1990abstracting}
Craig~A Knoblock.
\newblock Abstracting the tower of hanoi.
\newblock In \emph{Working Notes of AAAI-90 Workshop on Automatic Generation of
  Approximations and Abstractions}, pages 13--23. Citeseer, 1990.

\bibitem[Koenig and Simmons(1996)]{koenig1996effect}
Sven Koenig and Reid~G Simmons.
\newblock The effect of representation and knowledge on goal-directed
  exploration with reinforcement-learning algorithms.
\newblock \emph{Machine Learning}, 22\penalty0 (1):\penalty0 227--250, 1996.

\bibitem[Lekav{\`y} and N{\'a}vrat(2007)]{lekavy2007expressivity}
Mari{\'a}n Lekav{\`y} and Pavol N{\'a}vrat.
\newblock Expressivity of strips-like and htn-like planning.
\newblock In \emph{KES International Symposium on Agent and Multi-Agent
  Systems: Technologies and Applications}, pages 121--130. Springer, 2007.

\bibitem[Leonetti et~al.(2016)Leonetti, Iocchi, and
  Stone]{leonetti2016synthesis}
Matteo Leonetti, Luca Iocchi, and Peter Stone.
\newblock A synthesis of automated planning and reinforcement learning for
  efficient, robust decision-making.
\newblock \emph{Artificial Intelligence}, 241:\penalty0 103--130, 2016.

\bibitem[Matignon et~al.(2006)Matignon, Laurent, and
  Le~Fort-Piat]{matignon2006reward}
La{\"e}titia Matignon, Guillaume~J Laurent, and Nadine Le~Fort-Piat.
\newblock Reward function and initial values: Better choices for accelerated
  goal-directed reinforcement learning.
\newblock In \emph{International Conference on Artificial Neural Networks},
  pages 840--849. Springer, 2006.

\bibitem[McCallum(1995)]{mccallum1995reinforcement}
R~Andrew McCallum.
\newblock Reinforcement learning.
\newblock \emph{Advances in Neural Information Processing Systems 7},
  7:\penalty0 377, 1995.

\bibitem[McDermott(2000)]{mcdermott20001998}
Drew~M McDermott.
\newblock The 1998 ai planning systems competition.
\newblock \emph{AI magazine}, 21\penalty0 (2):\penalty0 35--35, 2000.

\bibitem[Rummery and Niranjan(1994)]{rummery1994line}
Gavin~A Rummery and Mahesan Niranjan.
\newblock \emph{On-line Q-learning using connectionist systems}, volume~37.
\newblock Citeseer, 1994.

\bibitem[Sacerdoti(1974)]{sacerdoti1974planning}
Earl~D Sacerdoti.
\newblock Planning in a hierarchy of abstraction spaces.
\newblock \emph{Artificial intelligence}, 5\penalty0 (2):\penalty0 115--135,
  1974.

\bibitem[Sacerdoti(1975)]{sacerdoti1975nonlinear}
Earl~D Sacerdoti.
\newblock The nonlinear nature of plans.
\newblock Technical report, STANFORD RESEARCH INST MENLO PARK CA, 1975.

\bibitem[Sussman(1973)]{sussman1973computational}
Gerald~J Sussman.
\newblock A computational model of skill acquisition.
\newblock 1973.

\bibitem[Sutton and Barto(1987)]{sutton1987temporal}
Richard~S Sutton and Andrew~G Barto.
\newblock A temporal-difference model of classical conditioning.
\newblock In \emph{Proceedings of the ninth annual conference of the cognitive
  science society}, pages 355--378. Seattle, WA, 1987.

\bibitem[Szepesv{\'a}ri and Littman(1996)]{szepesvari1996generalized}
Csaba Szepesv{\'a}ri and Michael~L Littman.
\newblock Generalized markov decision processes: Dynamic-programming and
  reinforcement-learning algorithms.
\newblock In \emph{Proceedings of International Conference of Machine
  Learning}, volume~96, 1996.

\bibitem[Taylor and Stone(2009)]{taylor2009transfer}
Matthew~E Taylor and Peter Stone.
\newblock Transfer learning for reinforcement learning domains: A survey.
\newblock \emph{Journal of Machine Learning Research}, 10\penalty0 (7), 2009.

\bibitem[Van~Otterlo and Wiering(2012)]{van2012reinforcement}
Martijn Van~Otterlo and Marco Wiering.
\newblock Reinforcement learning and markov decision processes.
\newblock In \emph{Reinforcement learning}, pages 3--42. Springer, 2012.

\bibitem[Van~Zanten(1990)]{van1990complexity}
AJ~Van~Zanten.
\newblock The complexity of an optimal algorithm for the generalized tower of
  hanoi problem.
\newblock \emph{International journal of computer mathematics}, 36\penalty0
  (1-2):\penalty0 1--8, 1990.

\bibitem[Vidyasagar(2020)]{vidyasagar2020recent}
M~Vidyasagar.
\newblock Recent advances in reinforcement learning.
\newblock In \emph{2020 American Control Conference (ACC)}, pages 4751--4756.
  IEEE, 2020.

\bibitem[Watkins and Dayan(1992)]{watkins1992q}
Christopher~JCH Watkins and Peter Dayan.
\newblock Q-learning.
\newblock \emph{Machine learning}, 8\penalty0 (3-4):\penalty0 279--292, 1992.

\bibitem[Watkins(1989)]{watkins1989learning}
Christopher John Cornish~Hellaby Watkins.
\newblock Learning from delayed rewards.
\newblock 1989.

\bibitem[White(1985)]{white1985real}
Douglas~J White.
\newblock Real applications of markov decision processes.
\newblock \emph{Interfaces}, 15\penalty0 (6):\penalty0 73--83, 1985.

\bibitem[Younes and Littman(2004)]{younes2004ppddl1}
H{\aa}kan~LS Younes and Michael~L Littman.
\newblock Ppddl1. 0: An extension to pddl for expressing planning domains with
  probabilistic effects.
\newblock \emph{Techn. Rep. CMU-CS-04-162}, 2:\penalty0 99, 2004.

\bibitem[Zimmerman and Kambhampati(2003)]{zimmerman2003learning}
Terry Zimmerman and Subbarao Kambhampati.
\newblock Learning-assisted automated planning: Looking back, taking stock,
  going forward.
\newblock \emph{AI Magazine}, 24\penalty0 (2):\penalty0 73--73, 2003.

\end{thebibliography}

\end{document}